\documentclass[12pt]{article}
\usepackage{arxiv}
\usepackage{amsmath}
\usepackage{graphicx,epstopdf}
\usepackage{amsmath}
\usepackage{times}
\usepackage{graphicx}
\usepackage{color}
\usepackage{multirow}
\usepackage{mathtools}
\usepackage{amssymb}
\usepackage{booktabs}       
\usepackage{amsfonts}       
\usepackage{nicefrac}       
\usepackage[authoryear]{natbib}	
\usepackage{tikz}
\usepackage{algorithm}
\usepackage{algpseudocode}
\title{A Dynamic Programming Algorithm for Finding an Optimal Sequence of Informative Measurements}

\date{}

\author{Peter N.~Loxley\\
University of New England.
\AND
Ka-Wai Cheung\\
University of New England.}

\renewcommand{\undertitle}{A Postprint}
\renewcommand{\headeright}{A Postprint}
\renewcommand{\shorttitle}{A Dynamic Programming Algorithm for Informative Measurements}

\begin{document}
\maketitle

\begin{abstract}
An informative measurement is the most efficient way to gain information about an unknown state. We present a first-principles derivation of a general-purpose dynamic programming algorithm that returns an optimal sequence of informative measurements by sequentially maximizing the entropy of possible measurement outcomes. This algorithm can be used by an autonomous agent or robot to decide where best to measure next, planning a path corresponding to an optimal sequence of informative measurements. The algorithm is applicable to states and controls that are either continuous or discrete, and agent dynamics that is either stochastic or deterministic; including Markov decision processes and Gaussian processes. Recent results from the fields of approximate dynamic programming and reinforcement learning, including on-line approximations such as rollout and Monte Carlo tree search, allow the measurement task to be solved in real time. The resulting solutions include non-myopic paths and measurement sequences that can generally outperform, sometimes substantially, commonly used greedy approaches. This is demonstrated for a global search task, where on-line planning for a sequence of local searches is found to reduce the number of measurements in the search by approximately half. A variant of the algorithm is derived for Gaussian processes for active sensing.
\end{abstract} 


\section{Introduction}\label{introduction}

Observing the outcomes of a sequence of measurements usually increases our knowledge about the state of a particular system we might be interested in. An informative measurement is the most efficient way of gaining this information, having the largest possible statistical dependence between the state being measured and the possible measurement outcomes. Lindley first introduced the notion of the amount of information in an experiment, and suggested the following greedy rule for experimentation: perform that experiment for which the expected gain in information is the greatest, and continue experimentation until a preassigned amount of information has been attained \citep{lind}. 

\begin{figure}[] 
\includegraphics[scale=1.2]{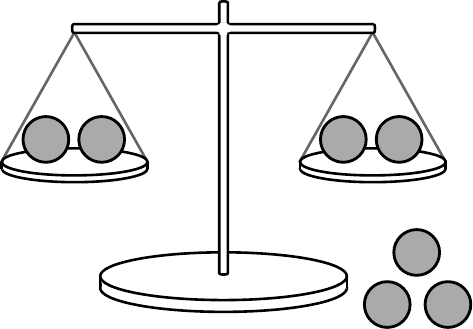}
\caption{A typical measurement task. In the weighing problem, an experimenter has a two-pan balance and a set of balls of equal weight except for a single odd ball that is either heavier or lighter than the others. The unknown state is the identity of the odd ball, and the possible measurement outcomes are ``left-pan heavier'', ``right-pan heavier'', or ``balanced'' if the odd ball is not on a pan. The objective is to find the odd ball in the fewest weighings.}
\label{fig1}
\end{figure}

Greedy methods are still the most common approaches for finding informative measurements, being both simple to implement and efficient to compute. Many of these approaches turn out to be Bayesian \citep{mackay,sivia}. For example, during Bayesian Adaptive \mbox{Exploration \citep{Loredo,Loredo2}}, measurement strategies are determined by maximizing the ``expected information'' for each ``observation-inference-design'' cycle. Alternatively, motivated by an inquiry-calculus still under development \citep{Kevin2}, a closely related method whereby the entropy of possible measurements is maximized for each cycle of ``inference'' and ``inquiry'' has demonstrated success across a diverse array of measurement tasks \citep{Kevin4,Kevin1,Placek_2017,Kevin3}. As a concrete example of a greedy measurement task, consider a weighing problem where an experimenter has a two-pan balance and is given a set of balls of equal weight except for a single odd ball that is either heavier or lighter than the others (see Figure \ref{fig1}). The experimenter would like to find the odd ball in the fewest weighings. MacKay suggested that for useful information to be gained as quickly as possible, each stage of an optimal measurement sequence should have measurement outcomes as close as possible to equiprobable \citep{mackay}. This is equivalent to choosing the measurement at each stage (i.e., the distribution of balls on pans) corresponding to the measurement outcome with largest entropy. 

Less well-recognized is the fact that greedy approaches can sometimes lead to suboptimal measurement sequences. This is not usually the case for simple systems such as the weighing problem described above. However, in other cases, an optimal sequence of measurements may involve trade-offs where earlier measurements in the sequence lead to modest information gains, rather than the maximum gains attainable at those stages, so that later measurements can access larger information gains than would otherwise be possible. A greedy approach never allows for such trade-offs. This is seen in the robotics literature, such as for path planning and trajectory optimization for robots building maps from sensor data \citep{placed,kollar}, path planning for active sensing \citep{low,cao}, and robots exploring unknown environments \citep{ay,kollar}. These approaches generally have in common the application of information theory objectives, and the use of sequential optimization methods such as dynamic programming or reinforcement learning \citep{bert,bert2,sutton}. Such methods specifically allow for the possibility of delayed information gains and non-myopic paths, and attempt to deal with the combinatorial nature of the underlying optimization problem. Exact dynamic programming is often computationally prohibitive or too time-consuming due to the need to consider all states of the problem. However, introducing efficient parametric approximations \citep{bert3, bert, loxley}, on-line approximations \citep{tes,bert,bert2}, or real-time heuristics \citep{barto}, allows very good suboptimal solutions to be found.

The aim of this work is to develop from first-principles a general-purpose dynamic programming algorithm for finding optimal sequences of informative measurements. We do not consider measurement problems with hidden states, such as hidden Markov models, that can only be observed indirectly through noisy measurements. This leads to a tractable algorithm that constructs an optimal sequence of informative measurements by sequentially maximizing the entropy of measurement outcomes. In addition to planning a sequence of measurements for a particular measurement problem, our algorithm can also be used by an autonomous agent or robot exploring a new environment to plan a path giving an optimal sequence of informative measurements. The framework we use for dynamic programming is very general, and includes Markov decision processes (MDPs) as a special case: allowing the agent dynamics to be either stochastic or deterministic, and the states and controls to be either continuous or discrete. Our general dynamic programming framework also allows suboptimal solution methods from approximate dynamic programming and reinforcement learning to be applied in a straightforward manner, avoiding the need to develop new approximation methods for each specific situation.

The closest related previous work is that of \citet{low} and \citet{cao} in the active-sensing domain. Active sensing is the task of deciding what data to gather next \citep{mackay92}, and is closely related to explorative behavior \citep{little_11,little_13}: for example, should an animal or agent gather more data to reduce its ignorance about a new environment, or should it omit gathering some data that are expected to be least informative. The work of \citet{low} and \citet{cao} models spatially-varying environmental phenomena using Gaussian processes \citep{rasmussen,gibbs,mackay}. Informative sampling paths are planned using various forms of dynamic programming. Gaussian processes are also described within our general dynamic programming framework. However, in the work of \citet{low} and \citet{cao}, it is not clear that dynamic programming is necessary for finding informative paths, as greedy algorithms are shown to achieve comparable results for the examples shown \citep{low,cao}. \citet{krause} showed that finding optimal static sensor placements for a Gaussian process requires solving a combinatorial optimization problem that is NP-complete \citep{krause}. Nevertheless, a greedy algorithm can achieve near-optimal performance when the information objective has the property of being submodular and monotonic \citep{krause,krause2,chen15b}. Further, for a particular path planning problem involving a Gaussian process, \citet{sing} presented an efficient recursive-greedy algorithm that attains near-optimal performance. In this work, we demonstrate a simple path planning problem that cannot be solved efficiently using a greedy algorithm. We also demonstrate the existence of delayed information gains in this example, indicating that solution methods must take into account trade-offs in information gains to find optimally informative paths. 

The structure of the paper is as follows. In Section \ref{sec2}, the general dynamic programming algorithm is developed from first-principles. The algorithm is then applied to two well-known examples (with slight modifications) in Section \ref{sec3} in order to illustrate the approach and present some exact solutions given by our algorithm. In Section \ref{sec4}, the algorithm is approximated by making use of approximate dynamic programming methodology to allow for real-time behavior of an autonomous agent or robot, and it is applied to a scaled-up on-line example where a greedy approach is shown to fail. In Section \ref{sec5}, a slight modification to the algorithm is given to describe Gaussian processes for active sensing and robot path planning. A conclusion is presented in Section \ref{sec6}.

\section{Sequential Maximization of Entropy}\label{sec2}

To determine the kinds of measurements that are most informative, we introduce two random variables $X$ and $M$, where $X$ describes the state of the system of interest and is defined over some suitable state-space, and $M$ describes the outcome of a measurement made on that system and is defined over the space of possible measurement outcomes. These random variables are dependent, because we hope a measurement outcome $M$ provides us with some information about the state $X$. This dependence can be described using the mutual information \citep{shannon}, given by
\begin{equation}
I(X;M)=H(X)-H(X|M),
\end{equation}
where the entropy $H(X)$ is the average uncertainty in the random variable $X$ describing the state, and the conditional entropy $H(X|M)$ is the average uncertainty in $X$ that remains after observing the outcome $M$ of a measurement performed on $X$. This means $I(X;M)$ is the average reduction in uncertainty about $X$ that results from knowledge of $M$ \citep{mackay}. Its minimum value, $I(X;M)=0$, is obtained when $X$ and $M$ are independent random variables, while its maximum value, $I(X;M)=H(X)$, is obtained when $H(X|M)=0$; so that if you know $M$ then you know everything there is to know about $X$. Given definitions for $X$ and $M$ corresponding to a particular system of interest, we would like to choose the measurement that maximizes $I(X;M)$ in order to reduce our uncertainty about $X$ and obtain the maximum amount of information available. 
\vspace{5pt}\\
\noindent
{\bf Assumption 1:}
\emph{
We now focus exclusively on measurement outcomes satisfying
\begin{equation}
H(M|X)=0.
\label{assump1}
\end{equation}
This means the measurement outcome $M$ of a state is fully determined (i.e., the uncertainty over $M$ is zero) given complete knowledge of the state $X$. For example, in the weighing problem shown in Figure \ref{fig1}, if the state of the odd ball is known with complete certainty to be $X=$ ``heavy ball on left pan'', then it is also known  that the measurement outcome will be $M=$ ``left-pan heavier''. Therefore, the only source of uncertainty in a measurement outcome is due to uncertainty in our knowledge of the state we are attempting to measure, rather than any measurement error. This is unlike the case of a hidden Markov model, or state-space model, where measurements (observations) of a state are noisy, and yield only partial state information. Here, we are modelling the case of complete state information rather than partial state information.}

The mutual information $I(X;M)$ can be written equivalently as $I(X;M)=$\linebreak$H(M)-H(M|X)$, so that Assumption 1 means $I(X;M)$ can be maximized over the probability distribution of measurement outcomes, $p_M(m)$, 
\begin{equation}
\underset{p_{(X,M)}}{\operatorname{max}}\ I(X;M)=\underset{p_M}{\operatorname{max}}\ H(M).
\end{equation} 
Another way of stating this is that $p_M(m)$ is a \emph{maximum entropy distribution} \citep{jaynes}. 

The entropy maximization will usually need to be done over a sequence of measurements, as it is unlikely that a single measurement will be sufficient to determine $X$ precisely when either a large number of states are present, or when measurement resolution is limited. For example, using a two-pan balance to find the odd ball in the weighing problem usually requires a sequence of weighings. Extending our approach from a single measurement with outcome $M$, to a sequence of measurements with outcomes $M_0,\ldots,M_{N-1}$, we look for a sequence of probability distributions $\{p_{M_0},\ldots, p_{M_{N-1}}\}$ that maximize the joint entropy $H(M_0,\ldots,M_{N-1})$. The key observation is that in many cases of interest this maximization can be carried out sequentially. Applying the chain rule for entropy \citep{cover}, leads to 
\begin{equation}
\underset{\{p_{M_0},\ldots,\hspace{1pt} p_{M_{N-1}}\}}{\operatorname{max}}\ \sum_{k=0}^{N-1}H_k(M_k|M_{k-1},\ldots,M_{0}),
\label{forgp}
\end{equation}
where $H_k(M_k|M_{k-1},\ldots,M_{0})$ becomes $H_0(M_{0})$ when $k=0$. It is now straightforward to see that if each $M_k$ can be modelled as an independent random variable, then we only need to find a sequence of probability distributions that maximize a sum of independent~entropies:
\begin{equation}
\underset{\{p_{M_0},\ldots,\hspace{1pt} p_{M_{N-1}}\}}{\operatorname{max}}\ \sum_{k=0}^{N-1}H_k(M_k),
\label{objective1}
\end{equation}
which is a much simpler task and can be done sequentially. The maximization in (\ref{objective1}) will be written as a dynamic program in the next section. Alternatively, if the $M_k$ are dependent random variables, sequential maximization of Equation (\ref{forgp}) can be done using the method of state augmentation, but comes with the cost of an enlarged state-space. This is demonstrated in Section 5 for the case of a Gaussian process.
 
\subsection{Dynamic Programming Algorithm}\label{alg}
Dynamic programming \citep{bert} is a general technique for carrying out sequential optimization. Provided the objective function can be decomposed as a sum over independent stages as in (\ref{objective1}), the principle of optimality guarantees that optimal solutions can be found using the technique of backward induction: that is, starting at the final stage of the problem (the tail subproblem) and sequentially working backwards towards the initial stage; at each stage using the solution of the previous subproblem to help find the solution to the current subproblem (for this reason, they are often called overlapping subproblems). The Equation~(\ref{objective1}) is close to ``the basic problem'' of dynamic programming (DP), and we will adopt the notation commonly used in DP. 

In order to maximize the sum of entropies in Equation (\ref{objective1}), we will need to introduce a set of parameters to maximize over. For this purpose, it turns out that two parameters for each measurement are sufficient (see the discussion below). Therefore, the probability distribution $p_M(m)$ is now assumed to depend on the two parameters $x$ and $u$ (note that $x$ is not related to the random variable $X$, but is standard notation in DP), giving $p_M(m)=p(m|x,u)$. For a sequence of $N$ independent measurements, the $k$th probability distribution then becomes,
\begin{equation}
p_{M_k}(m_k)=p_k(m_k|x_k,u_k),
\end{equation}
where $\{x_k\}_{k=0}^{N-1}$ and $\{u_k\}_{k=0}^{N-1}$ are sets of parameters allowing the sequence of probability distributions $\{p_0,\ldots, p_{N-1}\}$ to vary according to the measurement chosen at each stage of the sequence. Each parameter $u_k\in U_k(x_k)$ is chosen from the set of measurements $U_k(x_k)$ possible at stage $k$ and $x_k$, while each parameter $x_k\in S_k$ is then updated according to the discrete dynamical system:
\begin{equation}
x_{k+1}=f_k(x_k,u_k,m_k).
\label{mdynamics}
\end{equation}
In other words, the ``measurement state'' $x_{k+1}$ determines how the set of possible measurements changes from $U_k(x_k)$ to $U_{k+1}(x_{k+1})$ as a result of the measurement $u_k$ chosen, and the corresponding measurement outcome $m_k$ that is realized. Allowing the set of possible measurements to change at each stage of a measurement process in this way is a unique and defining feature of our model. To allow for \emph{closed-loop maximization} where this extra information can be used at each stage, we define a sequence of functions $\mu_k$ that map $x_k$ into $u_k=\mu_k(x_k)$. A \emph{policy} or a \emph{design} is then given by a sequence of these functions, one for each measurement:
$$\pi=\{\mu_0(x_0),\ldots,\mu_{N-1}(x_{N-1})\}.$$
Maximizing over policies, and adding a terminal entropy $H_N$ for stage $N$, allows \mbox{Equation (\ref{objective1})} to be written as
\begin{equation}
\underset{\pi}{\operatorname{max}}\ \sum_{k=0}^{N-1}H_k(M_k)+H_N.
\label{objective2}
\end{equation}
The final step is to write the objective in (\ref{objective2}) as an expectation. This can be done using the fact that entropy is the expected value of the Shannon information content: 
\begin{equation}
H(M)=\mathbb{E}\left(\log_2{\frac{1}{p(m)}}\right),
\end{equation}
where the expectation is taken over all measurement outcomes $m\in \mathrm{Im}\ M$, and where $\log_2{(1/p(m))}$ is the Shannon information content of measurement outcome $M=m$. Expressing the entropies in (\ref{objective2}) in terms of expectations, then using the linearity of expectation and the fact that $N$ is finite to interchange the summation and the expectation, now leads to an expression for the maximum expected value of $N$ information contents:
\begin{equation}
\underset{\pi}{\operatorname{max}}\ \mathbb{E}\left\{\sum_{k=0}^{N-1}h_k(x_k,\mu_k(x_k),m_k)+h_N(x_N)\right\},
\label{objective3}
\end{equation}
where the expectation is over $m_0,\ldots,m_{N-1}$, and the information content of the $k$th measurement is given by
\begin{equation}
h_k(x_k,\mu_k(x_k),m_k)=\log_2{\frac{1}{p_k(m_k|x_k,\mu_k(x_k))}}.
\end{equation}
\vspace{5pt}\\
\noindent
{\bf Proposition 1 (Dynamic programming algorithm):} 
\emph{The maximum entropy of $N$ measurements, as expressed by Equation (\ref{objective3}), is equal to $J_0(x_0)$ given by the last step of the following algorithm that starts with the terminal condition $J_N(x_N)=h_N(x_N)$, and proceeds backwards in time by evaluating the recurrence relation: 
\begin{equation}
J_k(x_k)=\underset{u_k\in U_k(x_k)}{\operatorname{max}}\ \underset{m_k}{\mathbb{E}}\Big\{h_k(x_k,u_k,m_k)+J_{k+1}(f_k(x_k,u_k,m_k))\Big\},
\label{dp}
\end{equation}
from the final stage $k=N-1$ to the initial stage $k=0$. The maximization in Equation (\ref{dp}) is over all~measurements $u_k\in U_k(x_k)$ possible at $x_k$ and stage $k$, while the expectation is over all~measurement outcomes $m_k\in \mathrm{Im}\ M_k$, and the function $h_k(x_k,u_k,m_k)$ is the information content of measurement outcome $M_k=m_k$:
\begin{equation}
h_k(x_k,u_k,m_k)=\log_2{\frac{1}{p_k(m_k|x_k,u_k)}}.
\label{infocontent}
\end{equation}
The optimal measurement sequence is given by the sequence $u_k^*=\mu_k^*(x_k)$ that maximizes the right~hand side of Equation (\ref{dp}) for each $x_k$ and $k$.}

The proof of this proposition is similar to that given in \citet{bert}. The procedure for the dynamic programming algorithm is outlined in Algorithm \ref{alg1}.
\begin{algorithm}[H]
\caption{Dynamic Programming Algorithm}\label{alg1}
\begin{algorithmic}
\State $J_N(x_N) \gets h_N(x_N)$
\vspace{2pt}
\For{$k=N-1$ \textbf{to} $0$}
\ForAll{$x_k\in S_k$}
\State $J_k(x_k)\gets \underset{u_k\in U_k(x_k)}{\operatorname{max}}\ \underset{m_k}{\mathbb{E}}\Big\{h_k(x_k,u_k,m_k)+J_{k+1}(f_k(x_k,u_k,m_k))\Big\}$
\vspace{2pt}
\State $\mu_k^*(x_k)\gets u_k^*$
\vspace{5pt}
\EndFor
\EndFor
\end{algorithmic}
\end{algorithm} 

There are two alternative ways to apply the DP algorithm that are both consistent with optimizing the objective in Equation (\ref{objective3}). For a fixed number of measurements $N$, the DP recurrence relation can be iterated over $N$ stages to determine the maximum information gained from $N$ measurements as given by $J_0$. Alternatively, a fixed amount of information can be gained over a minimum number of measurements by iterating the DP recurrence relation over a number of stages until we first reach this pre-assigned amount of information; whereupon we terminate the algorithm and read off the corresponding value of $N$. Are there smaller values of $N$ that would lead to this information gain? By construction we stopped the algorithm at the first stage we gained the required amount of information, so stopping any earlier would lead to a smaller information gain. We use both of these alternatives in Sections \ref{sec3} and \ref{sec4}.

\subsection{Extended Dynamic Programming Algorithm for an Autonomous Agent}\label{agentalg}
The previous DP algorithm allows us to find an optimal sequence of independent measurements by maximizing the entropy of $N$ independent measurement outcomes. We now include a simple extension to describe an autonomous agent seeking an optimal sequence of independent measurements as it explores a new environment. This opens up additional possibilities where dynamic programming can play a more substantial role. 

An agent moving through an environment is described by its position $x'_k\in S'_k$ at time $k$. The agent then decides to take control $u'_k\in U'_k(x'_k)$, moving it to a new position $x'_{k+1}$ at time $k+1$ according to the following dynamical system:
\begin{equation}
x'_{k+1}=v_k(x'_k,u'_k,w_k),
\label{agent}
\end{equation}

where $w_k\in D_k$ describes a random ``disturbance'' to the agent dynamics if the dynamics is stochastic. Coupling the agent dynamics to a sequence of measurements is achieved by augmenting the measurement state with the agent position, to give:
\begin{equation}
U_k(x_k)\mapsto U_k(x'_k,x_k),
\label{replacement}
\end{equation}

so the set of all measurements possible at stage $k$ now depends on the position of the agent in the environment at time $k$, as well as the measurement state during the $k$th measurement. The agent is assumed to take one measurement at each time step so the horizon of the agent dynamics is determined by the number of measurements in a sequence. It is possible to relax this assumption by introducing an extra index $k'$ that distinguishes the horizon of the agent dynamics from the number of measurements in a sequence, but we choose not to do this here. The DP recurrence relation given by (\ref{dp}) now becomes,
\begin{align}
J_k(x'_k,x_k)=
\underset{u'_k\in U'_k(x'_k)}{\operatorname{max}}\ \underset{w_k}{\mathbb{E}}\Big\{
\underset{u_k\in U_k(x'_k,x_k)}{\operatorname{max}}\ \underset{m_k}{\mathbb{E}}\Big\{h_k(x_k,u_k,m_k)\nonumber\\
+J_{k+1}(v_k(x'_k,u'_k,w_k), f_k(x_k,u_k,m_k))\Big\}\Big\},
\label{dpagent}
\end{align}
where $J_k(x'_k,x_k)$ now depends on both $x'_k$ and $x_k$ (due to state augmentation); and there is an additional expectation over $w_k$, and maximization over $u'_k$, in order to allow the agent to move from $x'_k$ to $x'_{k+1}$. The corresponding DP algorithm starts with $J_N(x'_N,x_N)$\linebreak$=h_N(x'_N,x_N)$, and proceeds backwards in time by evaluating the recurrence relation (\ref{dpagent}) from stage $k=N-1$ to stage $k=0$. This procedure is outlined in Algorithm \ref{alg2}. The last step of the algorithm returns $J_0(x'_0,x_0)$, the maximum entropy of $N$ measurements made by an autonomous agent. Given a choice of values for $x'_0$, then $J_0(x'_0,x_0)$ should also be maximized over $x'_0$ to give: $J_0(x_0)=\operatorname{max}_{x'_0}\ J_0(x'_0,x_0)$. The optimal measurement sequence is given by the sequence $u_k^*=\mu_k^*(x'_k,x_k)$ that jointly maximizes the right hand side of Equation (\ref{dpagent}) for each $x'_k$ and $x_k$ at each $k$, and the autonomous agent dynamics is determined by the sequence $u^{\prime *}_k=\mu^{\prime *}_k(x'_k)$ that maximizes the right hand side of Equation~(\ref{dpagent}) for each $x'_k$ at each $k$. 

\begin{algorithm}[H]
\caption{Extended Dynamic Programming Algorithm}\label{alg2}
\begin{algorithmic}
\State $J_N(x'_N,x_N) \gets h_N(x'_N,x_N)$
\vspace{3pt}
\For{$k=N-1$ \textbf{to} $0$}
\vspace{1pt}
\ForAll{$(x'_k,x_k)\in S'_k\times S_k$}
\State $J_k(x'_k,x_k)\gets
\underset{u'_k\in U'_k(x'_k)}{\operatorname{max}}\ \underset{w_k}{\mathbb{E}}\Big\{
\underset{u_k\in U_k(x'_k,x_k)}{\operatorname{max}}\ \underset{m_k}{\mathbb{E}}\Big\{h_k(x_k,u_k,m_k)$
\\ \hspace{90pt}
$+J_{k+1}(v_k(x'_k,u'_k,w_k), f_k(x_k,u_k,m_k))\Big\}\Big\}$
\State $\mu_k^*(x'_k,x_k)\gets u_k^*$
\vspace{8pt}
\State $\mu^{\prime *}_k(x'_k)\gets u^{\prime *}_k$
\vspace{6pt}
\EndFor
\EndFor
\end{algorithmic}
\end{algorithm}
\vspace{5pt}
{\bf Proposition 2 (Extended dynamic programming algorithm):}
\emph{The objective maximized by Algorithm \ref{alg2} is given by
\begin{equation}
\underset{\pi'}{\operatorname{max}}\ \mathbb{E'}\left\{
\underset{\pi}{\operatorname{max}}\ \mathbb{E}\left\{\sum_{k=0}^{N-1}h_k(x_k,\mu_k(x'_k,x_k),m_k)+h_N(x'_N,x_N)\right\}\right\},
\label{agentobjective}
\end{equation}
where the policies $\pi'=\{\mu'_0,\ldots,\mu'_{N-1}\}$ and $\pi=\{\mu_0,\ldots,\mu_{N-1}\}$ are sequences of functions given by
\begin{equation*}
\mu'_k(x'_k)=u'_k,\ \ \ \text{and}\ \ \ 
\mu_k(x'_k,x_k)=u_k,
\end{equation*}
and where $\mathbb{E}'$ is an expectation over $w_0,\ldots,w_{N-1}$, and $\mathbb{E}$ is an expectation over $m_0,\ldots,m_{N-1}$.}

The proof of Proposition 2 is given in Appendix 1. The extended DP algorithm in Algorithm \ref{alg2} looks computationally formidable due to the potentially large number of evaluations of (\ref{dpagent}) required. Fortunately, the constraint given by $u_k\in U(x'_k,x_k)$ will often limit the number of feasible measurement states $x_k$ for a given agent position $x'_k$, so that instead of $|S'_k\times S_k|$ potential evaluations of (\ref{dpagent}), there will be some smaller multiple of $|S'_k|$ evaluations required. 
 
\section{Illustrative Examples and Exact Solutions}\label{sec3}

In this section, we work through two textbook examples to illustrate the use of the proposed dynamic programming algorithm in the case of a simple measurement problem (a third example, ``Guess my number", is given in Appendix 2). The first example illustrates Algorithm \ref{alg1}, while the second example illustrates the extension given by Algorithm \ref{alg2}. Both examples have complete state information, discrete states and controls, and deterministic agent dynamics. However, \mbox{Algorithms \ref{alg1} and \ref{alg2}} can also be used when states and controls are continuous, or when the agent dynamics is stochastic. For example, in the case of continuous states and controls, it is often possible to use gradient methods to perform the maximization step in Algorithms \ref{alg1} and \ref{alg2}. All three~examples are taken from Chapter 4 of \citet{mackay}, with slight modification. 

\subsection{A Weighing Problem}

The first example is a slight variation of the weighing problem we considered in Figure~\ref{fig1}. In the general version of this problem, the unknown odd ball can be either heavy or light. Here, we simplify the problem so that the odd ball is always a heavy ball. The weighting problem is now this: given a set of balls of equal weight except for a single heavy ball, determine the minimum number of weighings of a two-pan balance that identifies the heavy ball. 

Let $X\in\{1,\ldots,n\}$ be the label of the heavy ball, and let $M$ be the outcome of a weighing, taking one of the values: ``left-pan heavier'', ``right-pan heavier'', or ``balanced''. If the outcome of a particular weighing is ``balanced'', then the heavy ball is one of the balls left off the two-pan balance. We also make the following definitions:
\begin{align*}
x_k &= \text{total number of balls to be weighed at stage } k,\\
u_k &= \text{number of balls on both balance pans at stage } k,
\end{align*}
--as well as assuming there are an equal number of balls on each pan so that $u_k$ is even (otherwise, a weighing experiment leads to a trivial result). If every ball is equally likely to be the heavy ball, then the following parameterizations hold:
\begin{align*}
p_k(m_k|x_k,u_k) &= 
\begin{cases}
u_k/2x_k & m_k= \text{``left-pan heavier'' or ``right-pan heavier''},\\
(x_k-u_k)/x_k & m_k = \text{``balanced''},
\end{cases}\\ 
f_k(x_k,u_k,m_k)&= \begin{cases}
u_k/2 & m_k= \text{``left-pan heavier'' or ``right-pan heavier''},\\
x_k-u_k & m_k = \text{``balanced''}.
\end{cases} 
\end{align*}
Here, $p_k(m_k|x_k,u_k)$ is simply the number of balls leading to measurement outcome $M=m_k$, divided by the total number of balls weighed at stage $k$. The number of balls to be weighed at the next stage, $x_{k+1}$, is then $f_k(x_k,u_k,m_k)=p_k(m_k|x_k,u_k) x_k$. With these definitions and parameterizations, Equations (\ref{dp}) and (\ref{infocontent}) lead to the DP recurrence relation:

\begin{align}
J_k(x_k)=\underset{u_k\in U^+_{\cal{E}}(x_k)}{\operatorname{max}}\left\{\frac{u_k}{x_k}\left (\log_2{\frac{2x_k}{u_k}}+J_{k+1}\left (\frac{u_k}{2}\right )\right )+\frac{x_k-u_k}{x_k}\left (\log_2{\frac{x_k}{x_k-u_k}}+J_{k+1}\left (x_k-u_k\right )\right )\right\},
\label{mp1}
\end{align}

where $U^{+}_{\cal{E}}(x_k)$ is the set $\{2,4,\ldots,x_k\}$ if $x_k$ is even, and $\{2,4,\ldots,x_k-1\}$ if $x_k$ is odd. 

Following the principle of optimality, the DP algorithm given by Algorithm \ref{alg1} starts at the final stage with terminal condition $J_N(1)=0$ bits, and proceeds backwards in time. From Equation (\ref{mp1}), the tail subproblem for measurement $k=N-1$ at $x_{N-1}=2$ becomes
\begin{align*}
J_{N-1}(2)&=\log_2{2}+J_{N}\left (1\right ),\\
&= 1\text{ bit.} \ \ \ (u^*_{N-1}=2)
\end{align*}
For $x_{N-1}=3$, the tail subproblem becomes, 
\begin{align*}
J_{N-1}(3)&=\frac{2}{3}\left (\log_2{3}+J_{N}\left (1\right )\right )+\frac{1}{3}\left (\log_2{3}+J_{N}\left (1\right )\right ),\\
&= \log_2{3}\text{ bits.} \ \ \ (u^*_{N-1}=2)
\end{align*}
The subproblem for $x_{k}=4$ now requires $J_{k+1}(2)$, according to Equation (\ref{mp1}). In this case, the tail subproblem given by $J_{N-1}(2)$ only becomes an overlapping subproblem if we move to measurement $k=N-2$, so that:  
\begin{align}
J_{N-2}(4)&=\operatorname{max}\left\{\frac{1}{2}\left (\log_2{4}+J_{N-1}\left (1\right )\right )+\frac{1}{2}\left (\log_2{2}+J_{N-1}\left (2\right )\right ),\ \ \log_2{2}+J_{N-1}\left (2\right )\right \},\nonumber\\
&= 2\text{ bits.} \ \ \ (u^*_{N-2}=2\text{ or }u^*_{N-2}=4)
\label{four}
\end{align}
We now have the exact DP solution to the weighing problem for four balls. The DP solution can be continued in this way by increasing both the number of balls and the number of measurements. As there are three states of the balance, and $n$ states for the $n$ possibilities where one of the balls is the heavy ball, the upper bound on the entropy (corresponding to equiprobable outcomes) is $\log_2{3}$ bits per weighing for $M$, and $\log_2{n}$ bits for $X$. Therefore, the heavy ball is guaranteed to be found after a number of weighings equal to $\lceil \log_2{n}/\log_2{3}\rceil$,  where the ceiling function $\lceil .\rceil$ rounds up to the closest integer.

The new contribution from DP can be seen in the two alternative solutions for $J_{N-2}(4)$ in (\ref{four}). In Solution 1 ($u^*_{N-2}=2$), we place one ball on each pan in the first weighting, and two balls are kept off the two-pan balance. With probability 0.5, we will ``get lucky'' by finding the heavy ball on the first weighing and immediately gain 2 bits of information. If not, then the heavy ball will be found on the second weighing. In addition to maximizing the entropy of two measurements, this solution also minimizes the average number of weighings to find the heavy ball. In Solution 2 ($u^*_{N-2}=4$), we place two balls on each pan in the first weighing, the outcome informing us which pair contains the heavy ball. The identity of the heavy ball is then resolved by placing each ball from this pair on a separate pan in the second weighing. There is no chance to ``get lucky'' in this case, since this solution always requires two weighings. 

Both of these solutions are optimal and maximize the entropy over two measurements (each giving 2 bits of information); however, Solution 1 has a larger entropy for the first measurement (1.5 bits versus 1 bit), while Solution 2 spreads the entropy more evenly between the two measurements. Therefore, seeking the most informative set of weighings by maximizing the entropy of \emph{each} measurement, as in \citet{mackay}, would lead only to Solution 1. Our DP algorithm finds all of the optimal solutions and therefore provides a more rigorous approach to solving this problem.

\subsection{Find the Submarine}\label{find-the-submarine}
\begin{figure}[H]
\begin{tikzpicture} [scale = 1.5]
\draw[step=0.5cm,color=gray] (-0.5,-0.5) grid (1,1);
\node at (0.25,0.75) {\huge $\times$};
\node at (-0.25,0.25) {\huge $\times$};
\node[fill=red] at (0.25,0.25) {\huge $\times$};
\node at (0.75,0.25) {\huge $\times$};
\node at (0.25,-0.25) {\huge $\times$};
\end{tikzpicture}\hspace{30pt}
\begin{tikzpicture} [scale = 1.5]
\draw[step=0.5cm,color=gray] (-0.5,-0.5) grid (1,1);
\node at (-0.25,0.75) {\large 1};
\node at (0.25,0.75) {\large 2};
\node at (0.75,0.75) {\large 3};
\node at (-0.25,0.25) {\large 4};
\node at (0.25,0.25) {\large 5};
\node at (0.75,0.25) {\large 6};
\node at (-0.25,-0.25) {\large 7};
\node at (0.25,-0.25) {\large 8};
\node at (0.75,-0.25) {\large 9};
\end{tikzpicture}
\caption{(\textbf{Left}) Sonar search pattern for a ship in ``Find the Submarine''. The red square is the position of the ship, and the ``$\times$'' symbols indicate the grid squares that are searched in a single measurement. (\textbf{Right}) Grid coordinates used for the position of the ship and the submarine.}
\label{fig2}
\end{figure}
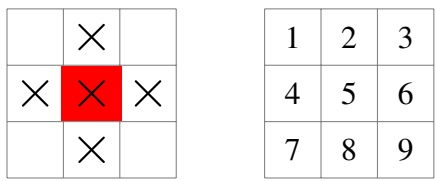

In the second example, a ship (treated as an autonomous agent) moves on a $3\times 3$ grid and uses its sonar to attempt to locate a stationary submarine positioned at a hidden location. To make the problem more interesting, we allow the sonar to search five neighbouring squares in a single measurement using the extended local search pattern shown in Figure \ref{fig2}. 

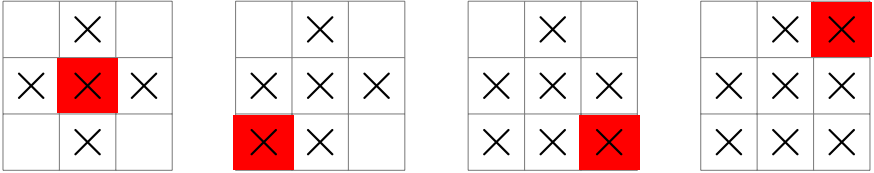
\begin{figure}[H]
\begin{tikzpicture} [scale = 1.5]
\draw[step=0.5cm,color=gray] (-0.5,-0.5) grid (1,1);
\node at (0.25,0.75) {\huge $\times$};
\node at (-0.25,0.25) {\huge $\times$};
\node[fill=red] at (0.25,0.25) {\huge $\times$};
\node at (0.75,0.25) {\huge $\times$};
\node at (0.25,-0.25) {\huge $\times$};
\end{tikzpicture}\hspace{20pt}
\begin{tikzpicture} [scale = 1.5]
\draw[step=0.5cm,color=gray] (-0.5,-0.5) grid (1,1);
\node at (0.25,0.75) {\huge $\times$};
\node at (-0.25,0.25) {\huge $\times$};
\node at (0.25,0.25) {\huge $\times$};
\node at (0.75,0.25) {\huge $\times$};
\node at (0.25,-0.25) {\huge $\times$};
\node[fill=red] at (-0.25,-0.25) {\huge $\times$};
\end{tikzpicture}\hspace{20pt}
\begin{tikzpicture} [scale = 1.5]
\draw[step=0.5cm,color=gray] (-0.5,-0.5) grid (1,1);
\node at (0.25,0.75) {\huge $\times$};
\node at (-0.25,0.25) {\huge $\times$};
\node at (0.25,0.25) {\huge $\times$};
\node at (0.75,0.25) {\huge $\times$};
\node at (0.25,-0.25) {\huge $\times$};
\node at (-0.25,-0.25) {\huge $\times$};
\node[fill=red] at (0.75,-0.25) {\huge $\times$};
\end{tikzpicture}\hspace{20pt}
\begin{tikzpicture} [scale = 1.5]
\draw[step=0.5cm,color=gray] (-0.5,-0.5) grid (1,1);
\node at (0.25,0.75) {\huge $\times$};
\node at (-0.25,0.25) {\huge $\times$};
\node at (0.25,0.25) {\huge $\times$};
\node at (0.75,0.25) {\huge $\times$};
\node at (0.25,-0.25) {\huge $\times$};
\node at (-0.25,-0.25) {\huge $\times$};
\node at (0.75,-0.25) {\huge $\times$};
\node[fill=red] at (0.75,0.75) {\huge $\times$};
\end{tikzpicture}
\caption{A search pattern initiated by a ship in the center of the grid. The measurement sequence starts at the left-most grid illustration at time $k=0$, and finishes at the right-most grid illustration at time $k=3$. The $u_k$ sequence is $5,1,1,1$ so that four measurements are guaranteed to locate the submarine. See text for details.}
\label{fig3}
\end{figure}

\begin{figure}[H]
\begin{tikzpicture} [scale = 1.5]
\draw[step=0.5cm,color=gray] (-0.5,-0.5) grid (1,1);
\node at (-0.25,0.75) {\huge $\times$};
\node[fill=red] at (-0.25,0.25) {\huge $\times$};
\node at (0.25,0.25) {\huge $\times$};
\node at (-0.25,-0.25) {\huge $\times$};
\end{tikzpicture}\hspace{20pt}
\begin{tikzpicture} [scale = 1.5]
\draw[step=0.5cm,color=gray] (-0.5,-0.5) grid (1,1);
\node at (-0.25,0.75) {\huge $\times$};
\node at (-0.25,0.25) {\huge $\times$};
\node at (0.25,0.25) {\huge $\times$};
\node at (-0.25,-0.25) {\huge $\times$};
\node at (0.75,0.75) {\huge $\times$};
\node[fill=red] at (0.75,0.25) {\huge $\times$};
\node at (0.75,-0.25) {\huge $\times$};
\end{tikzpicture}\hspace{20pt}
\begin{tikzpicture} [scale = 1.5]
\draw[step=0.5cm,color=gray] (-0.5,-0.5) grid (1,1);
\node at (-0.25,0.75) {\huge $\times$};
\node at (-0.25,0.25) {\huge $\times$};
\node at (0.25,0.25) {\huge $\times$};
\node at (-0.25,-0.25) {\huge $\times$};
\node at (0.75,0.75) {\huge $\times$};
\node at (0.75,0.25) {\huge $\times$};
\node at (0.75,-0.25) {\huge $\times$};
\node[fill=red] at (0.25,-0.25) {\huge $\times$};
\end{tikzpicture}
\caption{A search pattern initiated by a ship at the left edge of the grid. In this case, the $u_k$ sequence is $4,3,1$ so that three measurements are guaranteed to locate the submarine.}
\label{fig4}
\end{figure}
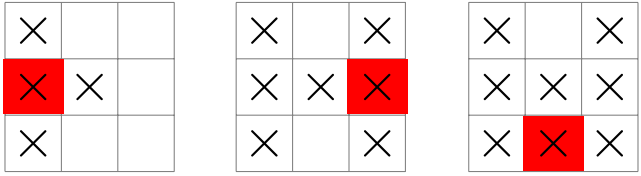

\begin{figure}[H]
\begin{tikzpicture} [scale = 1.5]
\draw[step=0.5cm,color=gray] (-0.5,-0.5) grid (1,1);
\node at (-0.25,0.25) {\huge $\times$};
\node[fill=red] at (-0.25,-0.25) {\huge $\times$};
\node at (0.25,-0.25) {\huge $\times$};
\end{tikzpicture}\hspace{20pt}
\begin{tikzpicture} [scale = 1.5]
\draw[step=0.5cm,color=gray] (-0.5,-0.5) grid (1,1);
\node at (-0.25,0.25) {\huge $\times$};
\node at (-0.25,-0.25) {\huge $\times$};
\node at (0.25,-0.25) {\huge $\times$};
\node[fill=red] at (0.75,-0.25) {\huge $\times$};
\node at (0.75,0.25) {\huge $\times$};
\end{tikzpicture}\hspace{20pt}
\begin{tikzpicture} [scale = 1.5]
\draw[step=0.5cm,color=gray] (-0.5,-0.5) grid (1,1);
\node at (-0.25,0.25) {\huge $\times$};
\node at (-0.25,-0.25) {\huge $\times$};
\node at (0.25,-0.25) {\huge $\times$};
\node at (0.75,-0.25) {\huge $\times$};
\node at (0.75,0.25) {\huge $\times$};
\node[fill=red] at (0.75,0.75) {\huge $\times$};
\node at (0.25,0.75) {\huge $\times$};
\end{tikzpicture}\hspace{20pt}
\begin{tikzpicture} [scale = 1.5]
\draw[step=0.5cm,color=gray] (-0.5,-0.5) grid (1,1);
\node at (-0.25,0.25) {\huge $\times$};
\node at (-0.25,-0.25) {\huge $\times$};
\node at (0.25,-0.25) {\huge $\times$};
\node at (0.75,-0.25) {\huge $\times$};
\node at (0.75,0.25) {\huge $\times$};
\node at (0.75,0.75) {\huge $\times$};
\node at (0.25,0.75) {\huge $\times$};
\node[fill=red] at (-0.25,0.75) {\huge $\times$};
\end{tikzpicture}
\caption{A search pattern initiated by a ship at the bottom-left corner of the grid. In this case, the $u_k$ sequence is $3,2,2,1$ so that four measurements are guaranteed to locate the submarine.}
\label{fig5}
\end{figure}
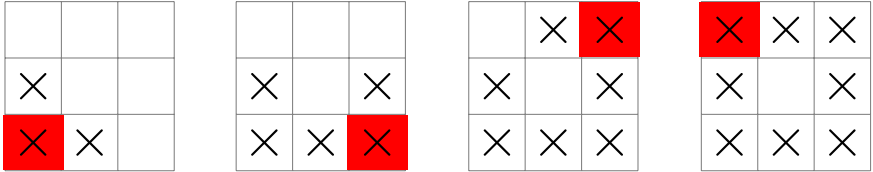

If the ship is located on a grid boundary, or if one of the neighbouring squares has already been searched in a previous measurement, then fewer than five squares will contribute new information to the current search (see Figures \ref{fig3}--\ref{fig5}). Further, if the submarine is located in any one of the five sonar search squares, we assume its precise location has been successfully determined. In these type of games, the ship can usually move to any square on the grid in one move. Instead, we choose a more realistic agent dynamics that~obeys
\begin{equation}
x'_{k+1}=x'_k+u'_k,
\label{agentdyn}
\end{equation}
where $x'_k$ is the position of the ship on the $3\times 3$ grid at time $k$, and $u'_k$ is a movement either along one of the Cartesian directions by two squares, or along a diagonal by one square. 

A reasonable approach to this problem might be to position the ship to search the largest possible area in the first measurement, corresponding to a greedy approach. Instead, the DP algorithm proposed here will maximize the entropy over the whole sequence of measurements, allowing for trade-offs in measurements over different stages. Since the agent dynamics given by (\ref{agentdyn}) is deterministic, the expectation over $w_k$ in Equation (\ref{dpagent}) vanishes. It is also the case that the maximum over $u_k$ in (\ref{dpagent}) is unnecessary in this example because there is no set of measurements to maximize over at each stage, only the sonar with its fixed search pattern. The DP recurrence relation (\ref{dpagent}) therefore simplifies to:

\begin{align}
J_k(x'_k,x_k)=
\underset{u'_k\in U'_k(x'_k)}{\operatorname{max}}\ \underset{m_k}{\mathbb{E}}\Big\{h_k(x_k,u_k(x'_k),m_k)+J_{k+1}(v_k(x'_k,u'_k), f_k(x_k,u_k(x'_k),m_k))\Big\}.
\label{dpmesagent}
\end{align}

Let the random variable $X$ be the position of the submarine on the $3\times 3$ grid using the grid coordinates shown in Figure \ref{fig2}, and let $M$ return ``yes'' or ``no'' to the question: ``Is the submarine detected by the sonar?''. If the answer is ``yes'', we assume the agent knows which particular square the submarine is located on, as previously mentioned. 
We then make the following definitions:
\begin{align*}
x_k &= \text{number of possible locations of submarine at stage } k,\\
u_k &= \text{number of new locations searched by sonar at stage } k,
\end{align*}
and the following parameterizations:
\begin{align*}
p_k(m_k|x_k,u_k) &= 
\begin{cases}
u_k/x_k & m_k= \text{``yes''},\\
(x_k-u_k)/x_k & m_k = \text{``no''},
\end{cases}\\ \\
f_k(x_k,u_k,m_k) &= 
\begin{cases}
u_k & m_k= \text{``yes''},\\
x_k-u_k & m_k = \text{``no''}.
\end{cases} 
\end{align*}
With these definitions and parameterizations, Equations  (\ref{infocontent}) and (\ref{dpmesagent}) lead to the following DP recurrence relation:
\begin{align}
J_k(x'_k,x_k)&=\underset{u'_k}{\operatorname{max}}\left\{\frac{u_k}{x_k}\left (\log_2{\frac{x_k}{u_k}}+J_{k+1}\left (x'_k+u'_k, u_k\right )\right )\right.\nonumber\\
&\left.\hspace{50pt} +\frac{x_k-u_k}{x_k}\left (\log_2{\frac{x_k}{x_k-u_k}}+J_{k+1}\left (x'_k+u'_k, x_k-u_k\right )\right )\right\},
\nonumber \\
&=\log_2{x_k}-\frac{x_k-u_k}{x_k}\big (\log_2{(x_k-u_k)}-\underset{u'_k}{\operatorname{max}}\ J_{k+1}\left (x'_k+u'_k, x_k-u_k\right )\big ),
\label{sub}
\end{align}
where $u_k=u_k(x'_k)\in \{0,1,\ldots,5\}$ depends on the location of the ship $x'_k$ at stage $k$ relative to the grid boundaries, and on how many new locations can be searched by the sonar from that location. In the second equation, the term $J_{k+1}(x'_k+u'_k, u_k)$ has been replaced with $\log_2{u_k}$ because an answer of ``yes'' implies the precise location of the submarine has been determined, immediately yielding $\log_2{u_k}$ bits of information.

Before applying the DP algorithm, let's consider some possible ship paths and search patterns. In Figure \ref{fig3}, the ship moves to the center square and searches the largest possible area with its first measurement at time $k=0$, giving $x'_0=5,x_0=9$, and $u_0=5$. If the submarine is not found, the ship then searches the remaining squares in the next time periods using further measurements. For example, at time $k=1$, the ship moves diagonally to the bottom-left corner, giving $x'_1=7,x_1=4$, and $u_1=1$. At time $k=2$, the ship moves two squares to the right, giving $x'_2=9,x_2=3$, and $u_2=1$. Furthermore, at time $k=3$, the ship moves two squares up, giving $x'_3=3,x_3=2$, and $u_3=1$. The position of the submarine is now guaranteed to be known in four measurements.

In the second case (shown in Figure \ref{fig4}), the ship moves to position $x'_0=4$ at time $k=0$ and searches four squares, giving $x_0=9$ and $u_0=4$. This choice allows it to search the remaining squares in fewer measurements than the first case, so that the position of the submarine is guaranteed to be known in three measurements instead of four.

In the third case (shown in Figure \ref{fig5}), the ship moves to position $x'_0=7$ at time $k=0$ and searches thee squares, giving $x_0=9$ and $u_0=3$. Compared with the other cases, the ship searches a smaller area in the first measurement. The ship then moves to each other~corner at later time periods, until completing the search in four measurements.

The DP algorithm given by Algorithm \ref{alg2} starts with the terminal condition $J_N(x'_N,x_N)$\linebreak$=0$ bits. Proceeding backwards in time, we evaluate (\ref{sub}) at $x_{N-1}$ for the tail subproblem at time $N-1$. Inspection of the figures indicates that the ship can be in any of the corners in cases 1 and 3, and at $x'_{N-1}\in\{2,4,6,8\}$ for case 2, due to the existing four-fold symmetry. In all cases, we have $x_{N-1}=2$ and $u_{N-1}(x'_{N-1})=1$ due to the geometry of the sonar pattern, and the positions of the unsearched squares, leading to
\begin{align*}
J_{N-1}(x'_{N-1},2)&=\log_2{2}-\frac{1}{2}\left (\log_2{1}-J_N(x'_N,x_N)\right),\\
&=1\ \text{bit}.
\end{align*}
The subproblem for time $N-2$ depends on each case. In case 1, $x_{N-2}=3$ and $u_{N-2}(x'_{N-2})$\linebreak$=1$. This leads to
\begin{align*}
J_{N-2}(x'_{N-2},3)&=\log_2{3}-\frac{2}{3}\big (\log_2{2}-\underset{u'_{N-2}}{\operatorname{max}}\ J_{N-1}\left (x'_{N-2}+u'_{N-2}, 2\right )\big),\\
&=\log_2{3}\ \text{bits},
\end{align*}
provided that:
\begin{equation}
x'_{N-1}=x'_{N-2}+u^{\prime *}_{N-2}.
\label{dyn1}
\end{equation}
In case 2, $x_{N-2}=5$ and $u_{N-2}(x'_{N-2})=3$. Similar reasoning leads to
\begin{align*}
J_{N-2}(x'_{N-2},5)&=\log_2{5}-\frac{2}{5}\big (\log_2{2}-\underset{u'_{N-2}}{\operatorname{max}}\ J_{N-1}\left (x'_{N-2}+u'_{N-2}, 2\right )\big),\\
&=\log_2{5}\ \text{bits}.
\end{align*}
In case 3, $x_{N-2}=4$ and $u_{N-2}(x'_{N-2})=2$. This leads to
\begin{align*}
J_{N-2}(x'_{N-2},4)&=\log_2{4}-\frac{1}{2}\big (\log_2{2}-\underset{u'_{N-2}}{\operatorname{max}}\ J_{N-1}\left (x'_{N-2}+u'_{N-2}, 2\right )\big),\\
&=\log_2{4}\ \text{bits}.
\end{align*}
We now solve the subproblem for time $N-3$. In case 1, $x_{N-3}=4$ and $u_{N-3}=1$, leading to
\begin{align*}
J_{N-3}(x'_{N-3},4)&=\log_2{4}-\frac{3}{4}\big (\log_2{3}-\underset{u'_{N-3}}{\operatorname{max}}\ J_{N-2}\left (x'_{N-3}+u'_{N-3}, 3\right )\big),\\
&=\log_2{4}\ \text{bits},
\end{align*}
provided that:
\begin{equation}
x'_{N-2}=x'_{N-3}+u^{\prime *}_{N-3}.
\label{dyn2}
\end{equation}
In case 2, $x_{N-3}=9$ and $u_{N-3}(x'_{N-3})=4$, leading to
\begin{align}
J_{N-3}(x'_{N-3},9)&=\log_2{9}-\frac{5}{9}\big (\log_2{5}-\underset{u'_{N-3}}{\operatorname{max}}\ J_{N-2}\left (x'_{N-3}+u'_{N-3}, 5\right )\big),\nonumber \\
&=\log_2{9}\ \text{bits}.
\label{subresult}
\end{align}
In case 3, $x_{N-3}=6$ and $u_{N-3}(x'_{N-2})=2$. This leads to
\begin{align*}
J_{N-3}(x'_{N-3},6)&=\log_2{6}-\frac{2}{3}\big (\log_2{4}-\underset{u'_{N-3}}{\operatorname{max}}\ J_{N-2}\left (x'_{N-3}+u'_{N-3}, 4\right )\big),\\
&=\log_2{6}\ \text{bits}.
\end{align*}
Since the total number of possible submarine locations is initially nine, and $x_{N-3}=9$ in Equation (\ref{subresult}), we can now terminate the algorithm and set $N=3$. 

The final step is to maximize $J_{0}(x'_{0},x_0)$ over $x'_0$. Comparing the values for $J_{0}(x'_{0},4)$, $J_{0}(x'_{0},6)$, and $J_{0}(x'_{0},9)$ yields $J_0(9)=\log_2{9}$ bits of information from three measurements, and $x'_0\in\{2,4,6,8\}$ for the initial position of the ship: each of these positions leads to $u_0(x'_0)=4$ due to the four-fold symmetry of the grid. The optimal ship movements are given by $x'_0$, and Equations (\ref{dyn1}) and (\ref{dyn2}) with the optimal controls in Table \ref{tab1}. 
\begin{table}
\centering
$\begin{array}{l|l|l|l}
x'_0&u^{\prime *}_0&u^{\prime *}_1\\
\hline
2&6&-4\text{ or }-2\\
4&2&-4\text{ or }2\\
6&-2&-2\text{ or }4\\
8&-6&2\text{ or }4
\end{array}$
\\\vspace{8pt}
\caption{Optimal controls for ship in ``Find the Submarine".}\label{tab1}
\end{table}
A greedy approach that maximizes the entropy of each measurement is equivalent to the suboptimal search pattern shown in Figure \ref{fig3}. A DP solution leading to an optimal search pattern is shown in Figure \ref{fig4}. These two figures bear a resemblance to the greedy and non-myopic schematics shown in Figure \ref{fig3} of \citet{bush}. This example will be scaled up and solved using approximate dynamic programming in the next section.

\section{Real-Time Approximate Dynamic Programming}\label{sec4}

The exact DP approach given in Algorithms \ref{alg1} and \ref{alg2} requires all states of the problem to be taken into account. For many problems, the number of states can be very large, and fast computation of an exact solution in real time is therefore not possible. A useful approximation that allows for real-time solutions is to look ahead one or more stages, simulate some possible paths going forwards in time all the way out to the horizon, then choose the next state from the simulated path with largest entropy. This is repeated at each stage. We do not need to consider more states than those actually visited during the simulation---a considerable saving when the number of states in the problem is large. This ``on-line'' approach leads to an efficient algorithm that approximates the problem, while hopefully also leading to good suboptimal solutions. For the special case of problems with deterministic dynamics, efficient algorithms already exist; including Dijkstra's shortest-path algorithm for discrete states \citep{bert} and an extension for continuous states \citep{tsit}, and the $A^*$ algorithm for discrete states \citep{hart1,hart2}. In the more general case of stochastic dynamics, the rollout algorithm \citep{tes,bert,bert2}, combined with adaptive Monte Carlo sampling techniques such as Monte Carlo Tree Search \citep{chang1,chang2}, leads to efficient algorithms. Other possibilities also include sequential Monte Carlo approaches \citep{zheng}. In this section, we develop an on-line algorithm for stochastic dynamics that allows for real-time behavior of an autonomous agent or path-planning robot seeking an optimal set of measurements as the measurement task is unfolding.

The first approximation is to restrict attention to limited lookahead. This can be done, for example, by introducing a one-step lookahead function $\tilde{J}_{k+1}$ that approximates the true function $J_{k+1}$. Denoting $\hat{J}_k$ as the general one-step lookahead approximation of $J_{k}$, we write the one-step lookahead approximation of Equation (\ref{dpagent}) as,

\begin{align}
\hat{J}_k(x'_k,x_k)=
\underset{u'_k\in U'_k(x'_k)}{\operatorname{max}}\ \underset{w_k}{\mathbb{E}}\Big\{
\underset{u_k\in U_k(x'_k,x_k)}{\operatorname{max}}\ \underset{m_k}{\mathbb{E}}\Big\{h_k(x_k,u_k,m_k)\nonumber\\
+\tilde{J}_{k+1}(v_k(x'_k,u'_k,w_k), f_k(x_k,u_k,m_k))\Big\}\Big\}.
\label{one-step-lookahead}
\end{align} 
The one-step lookahead function $\tilde{J}_{k+1}$ can be found using an on-line approximation, as we now describe. Given some \emph{{base policies}} (also called \emph{{base heuristics}}) $\{\hat{\mu}'_{k+1}(x'_{k+1}),\ldots,\hat{\mu}'_{N-1}(x'_{N-1})\}$ and $\{\hat{\mu}_{k+1}(x'_{k+1},x_{k+1}),\ldots,\hat{\mu}_{N-1}(x'_{N-1},x_{N-1})\}$, it is possible to simulate the dynamics using Equations (\ref{mdynamics}) and (\ref{agent}) from $k+1$ all the way to the horizon at $N-1$. This idea is used in Algorithm \ref{alg3} describing the stochastic rollout algorithm. During each stage of the rollout algorithm, simulation is used to find $\tilde{J}_{k+1}$ for each control $u'_k\in U'_k(x'_k)$, and each measurement $u_k\in U_k(x'_k,x_k)$ (lines 3--15). Following this, the values of $u'_k$ and $u_k$ that maximize the right-hand-side of Equation (\ref{one-step-lookahead}) are chosen, leading to the rollout policies $\bar{\mu}'_k(x'_k)$ and $\bar{\mu}_k(x'_k,x_k)$ (lines 16 and 17). Rollout policies are guaranteed to be no worse in performance than the base policies they are constructed from, at least for base policies that are \emph{{sequentially improving}} \citep{bert, bert2}. In practice, rollout policies are often found to perform dramatically better than this \citep{bert}.

To find $\tilde{J}_{k+1}$ for each pair $(u'_k,u_k)$, we use simulation and Monte Carlo sampling in Algorithm \ref{alg3}. Firstly, the samples $w_k$ and $m_k$ are drawn from the probability distributions for $W_k$ and $M_k$ in line 5, and used to simulate the dynamics of $(x'_k, x_k)$ for the control pair $(u'_k,u_k)$, to give $(x'_{k+1}, x_{k+1})$ in line 6. The rollout phase then takes place in lines 7--11, where $(x'_{k+1}, x_{k+1})$ is simulated by generating a pair of base policies, drawing samples for $w_i$ and $m_i$, and then applying Equations (\ref{mdynamics}) and (\ref{agent}), stage-by-stage until the horizon is reached. At each stage the information content $h_{i}(x_{i},\hat{\mu}_{i},m_{i})$ is collected, and added to the other stages to produce an estimate for $\tilde{J}_{k+1}(x'_{k+1}, x_{k+1})$. These steps are repeated many times, and the estimates for $h_{k}(x_{k},u_{k},m_{k})
+\tilde{J}_{k+1}(x'_{k+1}, x_{k+1})$ are then averaged to give $\tilde{Q}_{k}(x'_{k},x_{k},u'_k,u_k)$; where the expectations on Line 14 are approximated by their sample averages. 

\begin{algorithm}[H]
\caption{Stochastic Rollout Algorithm}\label{alg3}
\begin{algorithmic}[1]
\State \textbf{Input:} $(x'_0,x_0)\in S'_k\times S_k$
\vspace{4pt}
\For{$k=0$ \textbf{to} $N-1$}
\vspace{2pt}
\For{ \textbf{each} $(u'_k,u_k)\in U'_k(x'_k)\times U_k(x'_k,x_k)$}
\vspace{2pt}
\Repeat
\vspace{2pt}
\State $w_k\sim p_{W_k}$, \ \ \ $m_k\sim p_{M_k}$
\vspace{4pt}
\State $x'_{k+1}\gets v_k(x'_k,u'_k,w_k),\ \ \ x_{k+1}\gets f_k(x_k,u_k,m_k)$
\vspace{4pt}
\For{$i=k+1$ \textbf{to} $N-1$}
\vspace{4pt}
\State $\{\hat{\mu}'_{i}(x'_i),\hat{\mu}_{i}(x'_i,x_i)\}\gets\verb|Generate_base_policies|(x'_{i},x_{i})$
\vspace{4pt}
\State $w_i\sim p_{W_i}$, \ \ \ $m_i\sim p_{M_i}$
\vspace{4pt}
\State $x'_{i+1}\gets v_i(x'_i,\hat{\mu}'_i(x'_i),w_i),\ \ \ x_{i+1}\gets f_i(x_i,\hat{\mu}_i(x'_i,x_i),m_i)$
\vspace{6pt}
\EndFor
\vspace{2pt}
\State \texttt{Store:} $h_{k}(x_{k},u_{k},m_{k})
+\tilde{J}_{k+1}(x'_{k+1}, x_{k+1})$
\vspace{4pt}
\Until{a selected criterion is met}
\vspace{2pt}
\State $\tilde{Q}_{k}(x'_{k},x_{k},u'_k,u_k)\gets
\underset{w_{k}}{\mathbb{E}}\Big\{
\underset{m_{k}}{\mathbb{E}}\Big\{h_{k}(x_{k},u_{k},m_{k})
+\tilde{J}_{k+1}(x'_{k+1}, x_{k+1})\Big\}\Big\}$
\EndFor
\vspace{4pt}
\State $\hat{J}_k(x'_k,x_k)\gets
\underset{u'_k\in U'_k(x'_k)}{\operatorname{max}}\
\underset{u_k\in U_k(x'_k,x_k)}{\operatorname{max}} \tilde{Q}_{k}(x'_{k},x_{k},u'_k,u_k)$
\vspace{2pt}
\State $\bar{\mu}_k(x'_k,x_k)\gets u_k^*,\ \ \ \bar{\mu}^{\prime}_k(x'_k)\gets u^{\prime *}_k$
\vspace{4pt}
\State $w_k\sim p_{W_k}$, \ \ \ $m_k\sim p_{M_k}$
\vspace{4pt}
\State $x'_{k+1}\gets v_k(x'_k,\bar{\mu}'_k(x'_k),w_k),\ \ \ x_{k+1}\gets f_k(x_k,\bar{\mu}_k(x'_k,x_k),m_k)$
\vspace{6pt}
\EndFor
\end{algorithmic}
\end{algorithm}

There are several steps in Algorithm \ref{alg3} that can be made more efficient by using adaptive sampling methods such as Monte Carlo Tree Search. In line 3, some less worthwhile controls can either be sampled less often in lines 4--13, the simulation of those controls in lines 7--11 can be terminated early before reaching the horizon, or those controls may be discarded entirely. This can be done adaptively by using, for example, statistical tests or heuristics. There are also other options available, such as rolling horizons and terminal cost approximations. See \citet{bert} and references therein for a more complete discussion. 

Algorithm \ref{alg3} makes use of the subroutine \verb|Generate_base_policies|. For rollout to work, a base policy must be fast to evaluate. Here, we use the idea of multistep lookahead to generate base policies. Setting $\tilde{J}_{k+1}$ to zero in Equation (\ref{one-step-lookahead}), gives the zero-step lookahead~solution:
\begin{align}
\hat{J}_k(x'_k,x_k)=
\underset{u_k\in U_k(x'_k,x_k)}{\operatorname{max}}\ \underset{m_k}{\mathbb{E}}\Big\{h_k(x_k,u_k,m_k)\Big\},
\label{zerostep}
\end{align}
which corresponds to maximizing the entropy of the current measurement only. The next simplest choice is to approximate $\tilde{J}_{k+1}$ itself with a one-step lookahead:
\begin{align*}
\tilde{J}_{k+1}(x'_{k+1},x_{k+1})=
\underset{u'_{k+1}}{\operatorname{max}}\ \underset{w_{k+1}}{\mathbb{E}}\Big\{
\underset{u_{k+1}}{\operatorname{max}}\ \underset{m_{k+1}}{\mathbb{E}}\Big\{h_{k+1}(x_{k+1},u_{k+1},m_{k+1})\nonumber\\
+\tilde{J}_{k+2}(v_{k+1}(x'_{k+1},u'_{k+1},w_{k+1}), f_{k+1}(x_{k+1},u_{k+1},m_{k+1}))\Big\}\Big\},
\end{align*}
where $\tilde{J}_{k+2}$ is now an approximation of $J_{k+2}$. Setting $\tilde{J}_{k+2}$ to zero, leads to the following closed-form expression for one-step lookahead: 
\begin{align}
\hat{J}_k(x'_k,x_k)&=
\underset{u_{k}\in U_k(x'_k,x_k)}{\operatorname{max}}\ \underset{m_{k}}{\mathbb{E}}\Big\{
h_k(x_k,u_k,m_k)\nonumber\\
&+\underset{u'_k\in U'_k(x'_k)}{\operatorname{max}}\ \underset{w_k}{\mathbb{E}}\Big\{
\underset{u_{k+1}\in U_{k+1}(v_k,f_k)}{\operatorname{max}}\ \underset{m_{k+1}}{\mathbb{E}}\Big\{
h_{k+1}(f_k,u_{k+1},m_{k+1})\Big\}\Big\}\Big\}.
\label{one-step-result}
\end{align} 
This equation gives the first correction to the zero-step lookahead result (\ref{zerostep}), so that $\hat{J}_k$ now depends on the information content at $k$ and $k+1$. We now have a closed-form expression that depends on both $u'_k$ and $u_k$ (where $u'_k$ appears through $v_k(x'_k,u'_k,w_k)$ in the argument of $U_{k+1}$), so that Equation (\ref{one-step-result}) can be used to generate the base policies needed in Algorithm~\ref{alg3}. The subroutine is given in Algorithm \ref{alg4}. Instead of approximating the expectations in Equation (\ref{one-step-result}) by their sample averages, we apply an ``optimistic approach'' and use the single-sample estimates $w_k$, $m_k$, and $m_{k+1}$. The expression on Line 3 is a closed-form expression, so the maximizations leading to the control $u^{\prime *}_k$ and the measurements $u_k^*$ and $u_{k+1}^*$ can be done very quickly. Now $u_{k+1}^*$ is discarded (only the first stage is approximated for limited lookahead) to return a pair of base policies $\hat{\mu}'_k(x'_k)$ and $\hat{\mu}_k(x'_k,x_k)$. 

\begin{algorithm}[H]
\begin{algorithmic}[1]
\State \textbf{Input:} $x'_k,x_k$
\vspace{4pt}
\State $w_k\sim p_{W_k}, \ \ \ m_k\sim p_{M_k},\ \ \ m_{k+1}\sim p_{M_{k+1}}$
\vspace{4pt}
\State $\underset{u_{k}\in U_k(x'_k,x_k)}{\operatorname{max}}\Big\{
h_k(x_k,u_k,m_k)+\underset{u'_k\in U'_k(x'_k)}{\operatorname{max}}\
\underset{u_{k+1}\in U_{k+1}(v_k,f_k)}{\operatorname{max}}\Big\{
h_{k+1}(f_k(x_k,u_k,m_k),u_{k+1},m_{k+1})\Big\}\Big\}$
\vspace{8pt}
\State $\hat{\mu}_k(x'_k,x_k)\gets u_k^*,\ \ \ \hat{\mu}^{\prime}_k(x'_k)\gets u^{\prime *}_k$
\end{algorithmic}
\caption{Generate base policies (one possibility based on an optimistic one-step lookahead)}\label{alg4}
\end{algorithm}

The time efficiency of Algorithm \ref{alg3} strongly depends on how Monte Carlo sampling is performed. If it cannot be carried out within the time constraints of the real-time problem, then adaptive sampling techniques such as Monte Carlo tree search must be used. This may lead to some degradation in the quality of solutions, but the aim of these techniques is to reduce the risk of degradation while gaining substantial computational efficiencies. In some cases, the principle of certainty equivalence may hold for the agent dynamics and single-sample estimates may be sufficient for approximating expectations. In other cases, such as for Gaussian processes discussed in Section 5, a model for $p_{M_k}$ allows closed-form expressions for expectations instead of requiring expensive sampling techniques. In the limit of pure rollout (i.e., with no Monte Carlo sampling), the time complexity of Algorithm \ref{alg3} is ${\cal{O}}(N^2C)$; where $N$ is the number of measurements (or number of stages to reach the horizon), and $C=\underset{k}{\operatorname{max}}|U'_k\times U_k|$ is the maximum number of agent controls and measurement choices per stage. If $N$ is too large for real-time solutions, then further options are available from approximate dynamic programming and reinforcement learning, such as rolling horizons with terminal cost approximation, or various other forms of approximate policy iteration \citep{bert}. Algorithms \ref{alg3} and \ref{alg4} are now demonstrated using an example with deterministic agent dynamics.  

\subsection*{Find the Submarine on-Line}

In this section, we compare a greedy policy to one given by real-time approximate dynamic programming (Algorithms \ref{alg3} and \ref{alg4}) for the example ``Find the Submarine'' previously discussed. A greedy policy is the most appropriate comparison here, since any improvement in performance beyond a greedy policy will demonstrate planing in a real-time environment. The greedy policy is found to give optimal behavior up to a certain grid size, beyond which it completely fails. The approximate DP (rollout) policy continues to give (near) optimal performance for much larger grids, where planning is demonstrated to take place.

Algorithms \ref{alg3} and \ref{alg4} are appropriately modified to include a parametric model for $p_{M_k}$, deterministic agent dynamics, and only a single choice of measurement at each stage. In Algorithm \ref{alg3}, this means lines 4, 5, 9, 13, 18, and the expectation with respect to $w_{k}$ on line 14 are no longer required. Following from Equation (\ref{sub}) and (\ref{one-step-lookahead}) now becomes 
\begin{align}
\hat{J}_k(x'_k,x_k)=\log_2{x_k}-\frac{x_k-u_k}{x_k}\big (\log_2{(x_k-u_k)}-\underset{u'_k}{\operatorname{max}}\ \tilde{J}_{k+1}\left (x'_k+u'_k, x_k-u_k\right )\big ),
\label{notused}
\end{align}
where $u_k=u_k(x'_k)$. To generate base policies for use in the rollout algorithm, we use the same approach that led to equation (\ref{one-step-result}), yielding the closed-form expression:
\begin{align}
\hat{J}_k(x'_k,x_k)=\log_2{x_k}-\underset{u'_k}{\operatorname{min}}\ \frac{x_k-u_k-u_{k+1}}{x_k}\log_2{(x_k-u_k-u_{k+1})},
\label{sub-closedform}
\end{align}
where $u_k=u_k(x'_k)$, and $u_{k+1}=u_{k+1}(x'_k+u'_k)$. The minimization over $u'_k$ in Equation (\ref{sub-closedform}) is equivalent to maximizing the terms $u_k+u_{k+1}$. Therefore, instead of Equation (\ref{sub-closedform}), we equivalently have,
\begin{equation}
\hat{J}_k(x'_k)=u_k(x'_k)+\underset{u'_k}{\operatorname{max}}\ u_{k+1}(x'_k+u'_k).
\label{base-policy}
\end{equation}
The equation (\ref{base-policy}) can be derived from a DP algorithm with recurrence relation:
\begin{equation}
J_{k}(x'_{k})=u_k(x'_{k})+\underset{u'_k}{\operatorname{max}}\ J_{k+1}(x'_{k}+u'_{k}),
\label{simple-dp}
\end{equation}
that maximizes the objective: $\sum_{k=0}^{N-1} u_k$. A moment's reflection should convince the reader that this DP algorithm also solves ``Find the Submarine''. Therefore, instead of approximating Equation (\ref{sub}) to get  (\ref{notused}), we now approximate Equation (\ref{simple-dp}) to get
\begin{equation}
\hat{J}_{k}(x'_{k})=u_k(x'_{k})+\underset{u'_k}{\operatorname{max}}\ \tilde{J}_{k+1}(x'_{k}+u'_{k}),
\label{submarine-lookahead}
\end{equation}
where the base policy $\hat{\mu}'_k(x'_k)$ can be generated using Equation (\ref{base-policy}). Algorithms \ref{alg3} and \ref{alg4} can now be appropriately modified to suit Equations (\ref{base-policy}) and (\ref{submarine-lookahead}). During each stage of rollout, simulation is used to find $\tilde{J}_{k+1}$ for each control $u'_k\in U'_k(x'_k)$ taken at state $x'_k$, and the value of $u'_k$ that maximizes the right-hand-side of Equation (\ref{submarine-lookahead}) is chosen. This leads to the rollout policy $\bar{\mu}'_k(x'_k)$, and describes the path followed by the ship as it plans an optimal sequence of sonar measurements. 

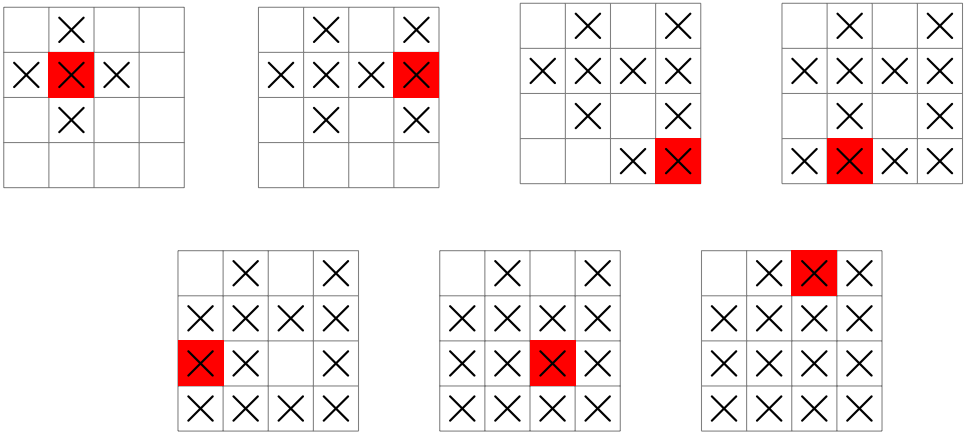
\begin{figure}[H]
\begin{tikzpicture} [scale=1.2]
\draw[step=0.5cm,color=gray] (0,0) grid (2,2);
\node at (0.25,1.25) {\huge $\times$};
\node at (0.75,1.75) {\huge $\times$};
\fill[red] (0.5,1.5) rectangle (1,1);
\node at (0.75,1.25) {\huge $\times$};
\node at (1.25,1.25) {\huge $\times$};
\node at (0.75,0.75) {\huge $\times$};
\end{tikzpicture}
\hspace{20pt}
\begin{tikzpicture} [scale=1.2]
\draw[step=0.5cm,color=gray] (0,0) grid (2,2);
\node at (0.25,1.25) {\huge $\times$};
\node at (0.75,1.75) {\huge $\times$};
\node at (0.75,1.25) {\huge $\times$};
\node at (1.25,1.25) {\huge $\times$};
\node at (0.75,0.75) {\huge $\times$};
\fill[red] (1.5,1.5) rectangle (2,1);
\node at (1.75,1.75) {\huge $\times$};
\node at (1.75,1.25) {\huge $\times$};
\node at (1.75,0.75) {\huge $\times$};
\end{tikzpicture}
\hspace{20pt}
\begin{tikzpicture} [scale=1.2]
\draw[step=0.5cm,color=gray] (0,0) grid (2,2);
\node at (0.25,1.25) {\huge $\times$};
\node at (0.75,1.75) {\huge $\times$};
\node at (0.75,1.25) {\huge $\times$};
\node at (1.25,1.25) {\huge $\times$};
\node at (0.75,0.75) {\huge $\times$};
\node at (1.75,1.75) {\huge $\times$};
\node at (1.75,1.25) {\huge $\times$};
\node at (1.75,0.75) {\huge $\times$};
\fill[red] (1.5,0.5) rectangle (2,0);
\node at (1.75,0.25) {\huge $\times$};
\node at (1.25,0.25) {\huge $\times$};
\end{tikzpicture}
\hspace{20pt}
\begin{tikzpicture} [scale=1.2]
\draw[step=0.5cm,color=gray] (0,0) grid (2,2);
\node at (0.25,1.25) {\huge $\times$};
\node at (0.75,1.75) {\huge $\times$};
\node at (0.75,1.25) {\huge $\times$};
\node at (1.25,1.25) {\huge $\times$};
\node at (0.75,0.75) {\huge $\times$};
\node at (1.75,1.75) {\huge $\times$};
\node at (1.75,1.25) {\huge $\times$};
\node at (1.75,0.75) {\huge $\times$};
\node at (1.75,0.25) {\huge $\times$};
\node at (1.25,0.25) {\huge $\times$};
\fill[red] (0.5,0.5) rectangle (1,0);
\node at (0.75,0.25) {\huge $\times$};
\node at (0.25,0.25) {\huge $\times$};
\end{tikzpicture}
\vspace{10pt}
\newline
\centering

\begin{tikzpicture} [scale=1.2]
\draw[step=0.5cm,color=gray] (0,0) grid (2,2);
\node at (0.25,1.25) {\huge $\times$};
\node at (0.75,1.75) {\huge $\times$};
\node at (0.75,1.25) {\huge $\times$};
\node at (1.25,1.25) {\huge $\times$};
\node at (0.75,0.75) {\huge $\times$};
\node at (1.75,1.75) {\huge $\times$};
\node at (1.75,1.25) {\huge $\times$};
\node at (1.75,0.75) {\huge $\times$};
\node at (1.75,0.25) {\huge $\times$};
\node at (1.25,0.25) {\huge $\times$};
\node at (0.75,0.25) {\huge $\times$};
\node at (0.25,0.25) {\huge $\times$};
\fill[red] (0,1) rectangle (0.5,0.5);
\node at (0.25,0.75) {\huge $\times$};
\end{tikzpicture}
\hspace{20pt}
\begin{tikzpicture} [scale=1.2]
\draw[step=0.5cm,color=gray] (0,0) grid (2,2);
\node at (0.25,1.25) {\huge $\times$};
\node at (0.75,1.75) {\huge $\times$};
\node at (0.75,1.25) {\huge $\times$};
\node at (1.25,1.25) {\huge $\times$};
\node at (0.75,0.75) {\huge $\times$};
\node at (1.75,1.75) {\huge $\times$};
\node at (1.75,1.25) {\huge $\times$};
\node at (1.75,0.75) {\huge $\times$};
\node at (1.75,0.25) {\huge $\times$};
\node at (1.25,0.25) {\huge $\times$};
\node at (0.75,0.25) {\huge $\times$};
\node at (0.25,0.25) {\huge $\times$};
\node at (0.25,0.75) {\huge $\times$};
\fill[red] (1,1) rectangle (1.5,0.5);
\node at (1.25,0.75) {\huge $\times$};
\end{tikzpicture}
\hspace{20pt}
\begin{tikzpicture} [scale=1.2]
\draw[step=0.5cm,color=gray] (0,0) grid (2,2);
\node at (0.25,1.25) {\huge $\times$};
\node at (0.75,1.75) {\huge $\times$};
\node at (0.75,1.25) {\huge $\times$};
\node at (1.25,1.25) {\huge $\times$};
\node at (0.75,0.75) {\huge $\times$};
\node at (1.75,1.75) {\huge $\times$};
\node at (1.75,1.25) {\huge $\times$};
\node at (1.75,0.75) {\huge $\times$};
\node at (1.75,0.25) {\huge $\times$};
\node at (1.25,0.25) {\huge $\times$};
\node at (0.75,0.25) {\huge $\times$};
\node at (0.25,0.25) {\huge $\times$};
\node at (0.25,0.75) {\huge $\times$};
\node at (1.25,0.75) {\huge $\times$};
\fill[red] (1,2) rectangle (1.5,1.5);
\node at (1.25,1.75) {\huge $\times$};
\end{tikzpicture}
\caption{A search pattern used by a ship following the greedy base policy for ``Find the Submarine''. The measurement sequence starts at the top-left grid illustration, then moves from left to right, top to bottom, before finishing at the bottom-right grid illustration. The $u_k$ sequence is $5,3,2,2,1,1,1$ so that seven measurements are guaranteed to locate the submarine.}
\label{fig6}
\end{figure}

The base policy generated using Algorithm \ref{alg4} with Equation (\ref{base-policy}) is shown in Figure \ref{fig6} for a $4\times 4$ grid.

This policy is greedy after the first stage: after the initial condition has been chosen, the policy seeks the maximum value of $u_k$ at each stage. Nevertheless, the greedy base policy turns out to be optimal for the $3\times3$ grid shown in Figures \ref{fig3}--\ref{fig5}, as well as for all grids up to $6\times 6$. 

\begin{figure}[H]
\begin{tikzpicture} [scale=1]
\draw[step=0.5cm,color=gray] (0,0) grid (3.5,3.5);
\fill[red] (3,3) rectangle (3.5,2.5);
\node at (3.25,2.75) {\huge $\times$};
\node at (3.25,2.25) {\huge $\times$};
\node at (3.25,3.25) {\huge $\times$};
\node at (2.75,2.75) {\huge $\times$};

\node at (0.25,3.25) {\huge $\times$};
\node at (0.75,3.25) {\huge $\times$};
\node at (1.25,3.25) {\huge $\times$};
\node at (2.25,3.25) {\huge $\times$};

\node at (0.75,2.75) {\huge $\times$};
\node at (1.75,2.75) {\huge $\times$};
\node at (2.25,2.75) {\huge $\times$};
\node at (2.75,2.75) {\huge $\times$};

\node at (0.25,2.25) {\huge $\times$};
\node at (0.75,2.25) {\huge $\times$};
\node at (1.25,2.25) {\huge $\times$};
\node at (1.75,2.25) {\huge $\times$};
\node at (2.25,2.25) {\huge $\times$};
\node at (2.75,2.25) {\huge $\times$};
\node at (3.25,2.25) {\huge $\times$};

\node at (0.25,1.75) {\huge $\times$};
\node at (0.75,1.75) {\huge $\times$};
\node at (1.25,1.75) {\huge $\times$};
\node at (1.75,1.75) {\huge $\times$};
\node at (2.25,1.75) {\huge $\times$};
\node at (2.75,1.75) {\huge $\times$};

\node at (0.25,1.25) {\huge $\times$};
\node at (0.75,1.25) {\huge $\times$};
\node at (1.25,1.25) {\huge $\times$};
\node at (1.75,1.25) {\huge $\times$};
\node at (2.25,1.25) {\huge $\times$};
\node at (2.75,1.25) {\huge $\times$};
\node at (3.25,1.25) {\huge $\times$};

\node at (0.25,0.75) {\huge $\times$};
\node at (0.75,0.75) {\huge $\times$};
\node at (1.25,0.75) {\huge $\times$};
\node at (1.75,0.75) {\huge $\times$};
\node at (2.25,0.75) {\huge $\times$};
\node at (2.75,0.75) {\huge $\times$};

\node at (0.25,0.25) {\huge $\times$};
\node at (0.75,0.25) {\huge $\times$};
\node at (1.25,0.25) {\huge $\times$};
\node at (1.75,0.25) {\huge $\times$};
\node at (2.25,0.25) {\huge $\times$};
\node at (2.75,0.25) {\huge $\times$};
\node at (3.25,0.25) {\huge $\times$};
\end{tikzpicture}
\hspace{20pt}
\begin{tikzpicture} [scale=1]
\draw[step=0.5cm,color=gray] (0,0) grid (3.5,3.5);
\node at (3.25,2.75) {\huge $\times$};
\node at (3.25,2.25) {\huge $\times$};
\node at (3.25,3.25) {\huge $\times$};
\node at (2.75,2.75) {\huge $\times$};

\node at (0.25,3.25) {\huge $\times$};
\node at (0.75,3.25) {\huge $\times$};
\node at (1.25,3.25) {\huge $\times$};
\node at (2.25,3.25) {\huge $\times$};

\fill[red] (3,2) rectangle (3.5,1.5);
\node at (3.25,1.75) {\huge $\times$};
\node at (0.75,2.75) {\huge $\times$};
\node at (1.75,2.75) {\huge $\times$};
\node at (2.25,2.75) {\huge $\times$};
\node at (2.75,2.75) {\huge $\times$};

\node at (0.25,2.25) {\huge $\times$};
\node at (0.75,2.25) {\huge $\times$};
\node at (1.25,2.25) {\huge $\times$};
\node at (1.75,2.25) {\huge $\times$};
\node at (2.25,2.25) {\huge $\times$};
\node at (2.75,2.25) {\huge $\times$};
\node at (3.25,2.25) {\huge $\times$};

\node at (0.25,1.75) {\huge $\times$};
\node at (0.75,1.75) {\huge $\times$};
\node at (1.25,1.75) {\huge $\times$};
\node at (1.75,1.75) {\huge $\times$};
\node at (2.25,1.75) {\huge $\times$};
\node at (2.75,1.75) {\huge $\times$};

\node at (0.25,1.25) {\huge $\times$};
\node at (0.75,1.25) {\huge $\times$};
\node at (1.25,1.25) {\huge $\times$};
\node at (1.75,1.25) {\huge $\times$};
\node at (2.25,1.25) {\huge $\times$};
\node at (2.75,1.25) {\huge $\times$};
\node at (3.25,1.25) {\huge $\times$};

\node at (0.25,0.75) {\huge $\times$};
\node at (0.75,0.75) {\huge $\times$};
\node at (1.25,0.75) {\huge $\times$};
\node at (1.75,0.75) {\huge $\times$};
\node at (2.25,0.75) {\huge $\times$};
\node at (2.75,0.75) {\huge $\times$};

\node at (0.25,0.25) {\huge $\times$};
\node at (0.75,0.25) {\huge $\times$};
\node at (1.25,0.25) {\huge $\times$};
\node at (1.75,0.25) {\huge $\times$};
\node at (2.25,0.25) {\huge $\times$};
\node at (2.75,0.25) {\huge $\times$};
\node at (3.25,0.25) {\huge $\times$};
\end{tikzpicture}
\hspace{20pt}
\begin{tikzpicture} [scale=1]
\draw[step=0.5cm,color=gray] (0,0) grid (3.5,3.5);
\node at (3.25,2.75) {\huge $\times$};
\node at (3.25,2.25) {\huge $\times$};
\node at (3.25,3.25) {\huge $\times$};
\node at (2.75,2.75) {\huge $\times$};

\node at (0.25,3.25) {\huge $\times$};
\node at (0.75,3.25) {\huge $\times$};
\node at (1.25,3.25) {\huge $\times$};
\node at (2.25,3.25) {\huge $\times$};

\node at (3.25,1.75) {\huge $\times$};
\node at (0.75,2.75) {\huge $\times$};
\node at (1.75,2.75) {\huge $\times$};
\node at (2.25,2.75) {\huge $\times$};
\node at (2.75,2.75) {\huge $\times$};

\node at (0.25,2.25) {\huge $\times$};
\node at (0.75,2.25) {\huge $\times$};
\node at (1.25,2.25) {\huge $\times$};
\node at (1.75,2.25) {\huge $\times$};
\node at (2.25,2.25) {\huge $\times$};
\node at (2.75,2.25) {\huge $\times$};
\node at (3.25,2.25) {\huge $\times$};

\fill[red] (3,1) rectangle (3.5,0.5);
\node at (3.25,0.75) {\huge $\times$};
\node at (0.25,1.75) {\huge $\times$};
\node at (0.75,1.75) {\huge $\times$};
\node at (1.25,1.75) {\huge $\times$};
\node at (1.75,1.75) {\huge $\times$};
\node at (2.25,1.75) {\huge $\times$};

\node at (3.25,1.75) {\huge $\times$};
\node at (2.75,1.75) {\huge $\times$};
\node at (0.25,1.25) {\huge $\times$};
\node at (0.75,1.25) {\huge $\times$};
\node at (1.25,1.25) {\huge $\times$};
\node at (1.75,1.25) {\huge $\times$};
\node at (2.25,1.25) {\huge $\times$};
\node at (2.75,1.25) {\huge $\times$};
\node at (3.25,1.25) {\huge $\times$};

\node at (0.25,0.75) {\huge $\times$};
\node at (0.75,0.75) {\huge $\times$};
\node at (1.25,0.75) {\huge $\times$};
\node at (1.75,0.75) {\huge $\times$};
\node at (2.25,0.75) {\huge $\times$};
\node at (2.75,0.75) {\huge $\times$};

\node at (0.25,0.25) {\huge $\times$};
\node at (0.75,0.25) {\huge $\times$};
\node at (1.25,0.25) {\huge $\times$};
\node at (1.75,0.25) {\huge $\times$};
\node at (2.25,0.25) {\huge $\times$};
\node at (2.75,0.25) {\huge $\times$};
\node at (3.25,0.25) {\huge $\times$};
\end{tikzpicture}
\caption{A ship following the greedy base policy moves away from a region of unsearched squares during its search (grids left to right). In the right-most grid illustration, all controls now equally maximize the next-stage entropy from this position, so a return to the region of unsearched squares is not guaranteed (for example, the ship can move horizontally, forwards and backwards in an endless cycle, without encountering an unsearched square). }
\label{fig7}
\end{figure}
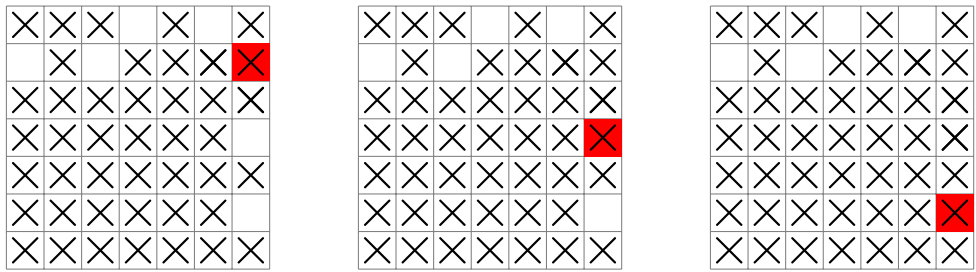

For grids larger than  $6\times 6$, the greedy base policy no longer works, and it becomes necessary to plan each measurement to obtain an optimal search pattern. The reason can be seen in Figure \ref{fig7}, which shows a ship following the greedy base policy on a $7\times 7$ grid.

The grid is now large enough that a ship can move out of reach of a region of unsearched squares, as shown in Figure \ref{fig7}. This is not possible for smaller grids such as those in Figure \ref{fig6}, because any unsearched squares will always be within reach of an admissible control (i.e., a control satisfying $u'_k\in U'_k(x'_k)$). As the grid gets larger, however, moving the ship back to a region of unsearched squares becomes more and more difficult under the greedy base policy, since it becomes possible for \emph{all} controls to maximize the entropy of the next stage (see Figure \ref{fig7}). In this case, the control that is chosen no longer depends on the entropy of a measurement, but rather, on the order the admissible controls are processed in.   

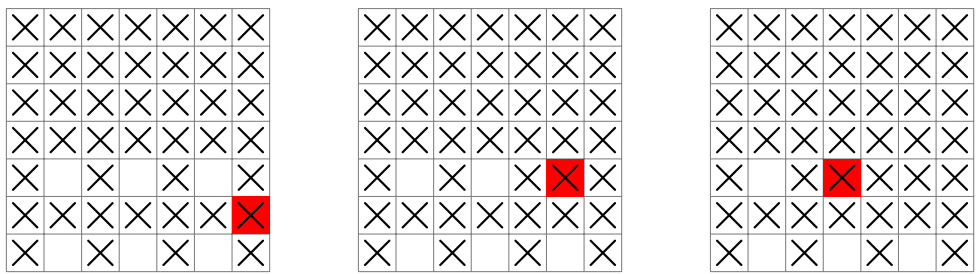
\begin{figure}[H]
\begin{tikzpicture} [scale=1]
\draw[step=0.5cm,color=gray] (0,0) grid (3.5,3.5);
\node at (0.25,3.25) {\huge $\times$};
\node at (0.75,3.25) {\huge $\times$};
\node at (1.25,3.25) {\huge $\times$};
\node at (1.75,3.25) {\huge $\times$};
\node at (2.25,3.25) {\huge $\times$};
\node at (2.75,3.25) {\huge $\times$};
\node at (3.25,3.25) {\huge $\times$};

\node at (0.25,2.75) {\huge $\times$};
\node at (0.75,2.75) {\huge $\times$};
\node at (1.25,2.75) {\huge $\times$};
\node at (1.75,2.75) {\huge $\times$};
\node at (2.25,2.75) {\huge $\times$};
\node at (2.75,2.75) {\huge $\times$};
\node at (3.25,2.75) {\huge $\times$};

\node at (0.25,2.25) {\huge $\times$};
\node at (0.75,2.25) {\huge $\times$};
\node at (1.25,2.25) {\huge $\times$};
\node at (1.75,2.25) {\huge $\times$};
\node at (2.25,2.25) {\huge $\times$};
\node at (2.75,2.25) {\huge $\times$};
\node at (3.25,2.25) {\huge $\times$};

\node at (0.25,1.75) {\huge $\times$};
\node at (0.75,1.75) {\huge $\times$};
\node at (1.25,1.75) {\huge $\times$};
\node at (1.75,1.75) {\huge $\times$};
\node at (2.25,1.75) {\huge $\times$};
\node at (2.75,1.75) {\huge $\times$};
\node at (3.25,1.75) {\huge $\times$};

\node at (0.25,1.25) {\huge $\times$};
\node at (1.25,1.25) {\huge $\times$};
\node at (2.25,1.25) {\huge $\times$};
\node at (3.25,1.25) {\huge $\times$};

\fill[red] (3,1) rectangle (3.5,0.5);
\node at (0.25,0.75) {\huge $\times$};
\node at (0.75,0.75) {\huge $\times$};
\node at (1.25,0.75) {\huge $\times$};
\node at (1.75,0.75) {\huge $\times$};
\node at (2.25,0.75) {\huge $\times$};
\node at (2.75,0.75) {\huge $\times$};
\node at (3.25,0.75) {\huge $\times$};

\node at (0.25,0.25) {\huge $\times$};
\node at (1.25,0.25) {\huge $\times$};
\node at (2.25,0.25) {\huge $\times$};
\node at (3.25,0.25) {\huge $\times$};
\end{tikzpicture}
\hspace{20pt}
\begin{tikzpicture} [scale=1]
\draw[step=0.5cm,color=gray] (0,0) grid (3.5,3.5);
\node at (0.25,3.25) {\huge $\times$};
\node at (0.75,3.25) {\huge $\times$};
\node at (1.25,3.25) {\huge $\times$};
\node at (1.75,3.25) {\huge $\times$};
\node at (2.25,3.25) {\huge $\times$};
\node at (2.75,3.25) {\huge $\times$};
\node at (3.25,3.25) {\huge $\times$};

\node at (0.25,2.75) {\huge $\times$};
\node at (0.75,2.75) {\huge $\times$};
\node at (1.25,2.75) {\huge $\times$};
\node at (1.75,2.75) {\huge $\times$};
\node at (2.25,2.75) {\huge $\times$};
\node at (2.75,2.75) {\huge $\times$};
\node at (3.25,2.75) {\huge $\times$};

\node at (0.25,2.25) {\huge $\times$};
\node at (0.75,2.25) {\huge $\times$};
\node at (1.25,2.25) {\huge $\times$};
\node at (1.75,2.25) {\huge $\times$};
\node at (2.25,2.25) {\huge $\times$};
\node at (2.75,2.25) {\huge $\times$};
\node at (3.25,2.25) {\huge $\times$};

\node at (0.25,1.75) {\huge $\times$};
\node at (0.75,1.75) {\huge $\times$};
\node at (1.25,1.75) {\huge $\times$};
\node at (1.75,1.75) {\huge $\times$};
\node at (2.25,1.75) {\huge $\times$};
\node at (2.75,1.75) {\huge $\times$};
\node at (3.25,1.75) {\huge $\times$};

\fill[red] (2.5,1.5) rectangle (3,1);
\node at (0.25,1.25) {\huge $\times$};
\node at (1.25,1.25) {\huge $\times$};
\node at (2.25,1.25) {\huge $\times$};
\node at (2.75,1.25) {\huge $\times$};
\node at (3.25,1.25) {\huge $\times$};

\node at (0.25,0.75) {\huge $\times$};
\node at (0.75,0.75) {\huge $\times$};
\node at (1.25,0.75) {\huge $\times$};
\node at (1.75,0.75) {\huge $\times$};
\node at (2.25,0.75) {\huge $\times$};
\node at (2.75,0.75) {\huge $\times$};
\node at (3.25,0.75) {\huge $\times$};

\node at (0.25,0.25) {\huge $\times$};
\node at (1.25,0.25) {\huge $\times$};
\node at (2.25,0.25) {\huge $\times$};
\node at (3.25,0.25) {\huge $\times$};
\end{tikzpicture}
\hspace{20pt}
\begin{tikzpicture} [scale=1]
\draw[step=0.5cm,color=gray] (0,0) grid (3.5,3.5);
\node at (0.25,3.25) {\huge $\times$};
\node at (0.75,3.25) {\huge $\times$};
\node at (1.25,3.25) {\huge $\times$};
\node at (1.75,3.25) {\huge $\times$};
\node at (2.25,3.25) {\huge $\times$};
\node at (2.75,3.25) {\huge $\times$};
\node at (3.25,3.25) {\huge $\times$};

\node at (0.25,2.75) {\huge $\times$};
\node at (0.75,2.75) {\huge $\times$};
\node at (1.25,2.75) {\huge $\times$};
\node at (1.75,2.75) {\huge $\times$};
\node at (2.25,2.75) {\huge $\times$};
\node at (2.75,2.75) {\huge $\times$};
\node at (3.25,2.75) {\huge $\times$};

\node at (0.25,2.25) {\huge $\times$};
\node at (0.75,2.25) {\huge $\times$};
\node at (1.25,2.25) {\huge $\times$};
\node at (1.75,2.25) {\huge $\times$};
\node at (2.25,2.25) {\huge $\times$};
\node at (2.75,2.25) {\huge $\times$};
\node at (3.25,2.25) {\huge $\times$};

\node at (0.25,1.75) {\huge $\times$};
\node at (0.75,1.75) {\huge $\times$};
\node at (1.25,1.75) {\huge $\times$};
\node at (1.75,1.75) {\huge $\times$};
\node at (2.25,1.75) {\huge $\times$};
\node at (2.75,1.75) {\huge $\times$};
\node at (3.25,1.75) {\huge $\times$};

\fill[red] (1.5,1.5) rectangle (2,1);
\node at (0.25,1.25) {\huge $\times$};
\node at (1.25,1.25) {\huge $\times$};
\node at (1.75,1.25) {\huge $\times$};
\node at (2.25,1.25) {\huge $\times$};
\node at (2.75,1.25) {\huge $\times$};
\node at (3.25,1.25) {\huge $\times$};

\node at (0.25,0.75) {\huge $\times$};
\node at (0.75,0.75) {\huge $\times$};
\node at (1.25,0.75) {\huge $\times$};
\node at (1.75,0.75) {\huge $\times$};
\node at (2.25,0.75) {\huge $\times$};
\node at (2.75,0.75) {\huge $\times$};
\node at (3.25,0.75) {\huge $\times$};

\node at (0.25,0.25) {\huge $\times$};
\node at (1.25,0.25) {\huge $\times$};
\node at (2.25,0.25) {\huge $\times$};
\node at (3.25,0.25) {\huge $\times$};
\end{tikzpicture}
\caption{A ship following the rollout policy conducts its search systematically (grid illustrations left to right), following a planned sequence of moves that avoids the ship moving out of reach of a region of unsearched squares.}
\label{fig8}
\end{figure}

Surprisingly, although the greedy base policy does not exhibit optimal behavior for larger grids, it can be used to improve a policy that subsequently attains optimal behavior. This improved policy (the rollout policy) is used to plan an optimal search pattern. A ship following the rollout policy is shown in Figure \ref{fig8} for a $7\times 7$ grid. 

During the late stages of this policy, the ship searches the remaining squares systematically by following a planned sequence of moves. Regardless of the initial condition, the planning done by the rollout policy avoids the ship moving out of reach of a region of unsearched squares as it does in Figure \ref{fig7}. As the grid increases in size, this behavior continues, and the minimum number of measurements guaranteed to find the submarine is shown in Table \ref{tab2}. 

\begin{table}
\centering
\begin{tabular}{c|c|c|c}
Grid Size & Squares & Number of Measurements&Percentage\\
\hline
\hline
$7\times 7$&49&23&47.9\\
$8\times 8$& 64&31&49.2\\
$9\times 9$&81&39&48.8\\
$10\times 10$&100&49&49.5\\
$11\times 11$&121&60&50.0\\
$12\times 12$&144&71&49.7\\
$13\times 13$& 169&84&50.0\\
$14\times 14$&196&98&50.3
\end{tabular}
\\\vspace{10pt}
\caption{The minimum number of measurements guaranteed to find the submarine using the rollout policy, shown for different grid sizes. Percentage is given by number of measurements divided by total number of squares searched.}\label{tab2}
\end{table}

In order to guarantee finding the submarine, it is necessary to search each square of the grid except for the last square (if we have not already found the submarine, it is guaranteed to be on the last square). However, according to the results in Table \ref{tab2},  the minimum number of measurements is approximately half the total number of squares that are searched. In other words, using a local search pattern such as the one shown in Figure~\ref{fig2} can substantially reduce the number of measurements taken to find the submarine by using dynamic programming to plan a path that includes a (near) optimal measurement sequence. 

\begin{figure}[H]
\includegraphics[scale=0.4]{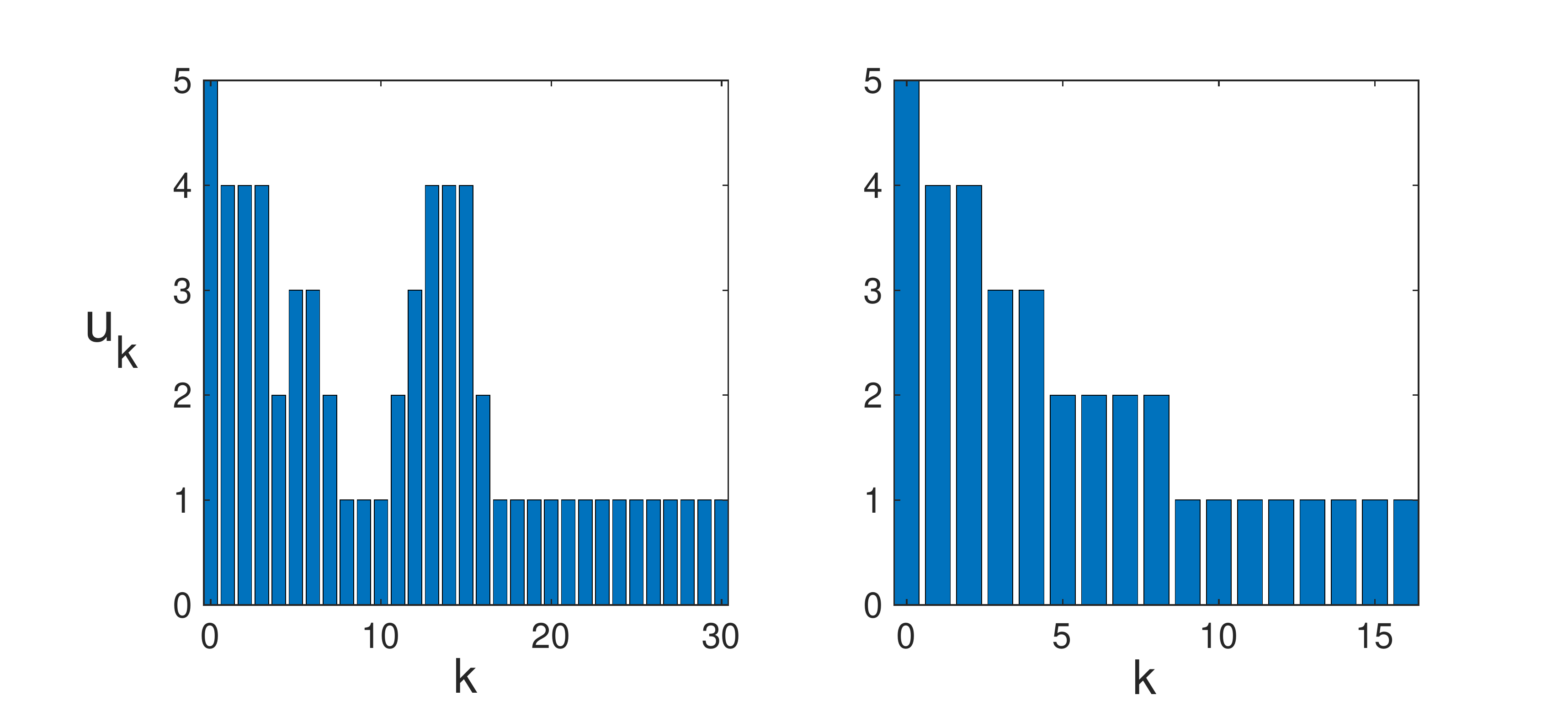}
\caption{(\textbf{Left}) Bar graph of $u_k$ versus $k$ for the rollout policy on an $8\times 8$ grid with 31 measurements. The two peaks at $k=5$ and $k=14$ demonstrate delayed information gains due to planned trade-offs where early low-entropy measurements lead to later higher-entropy measurements. (\textbf{Right}) Bar graph of $u_k$ versus $k$ for the greedy base policy on a $6\times 6$ grid with 17 measurements. In this case, there are no planning trade-offs, and $u_k$ is a strictly decreasing function of $k$.}
\label{fig9}
\end{figure}

In Figure \ref{fig9}, the time series of $u_k$ is shown for both the rollout and greedy base policies. 

Trade-offs leading to delayed information gains can clearly be seen for the rollout policy, where early measurements with lower entropy lead to later measurements with higher entropy, but not for the greedy base policy. These trade-offs are a sign of planning taking place in the rollout policy. It is interesting to note that while maximizing entropy always leads to a uniform probability distribution in the absence of external constraints, maximizing the entropy sequentially generally leads to a sequence of non-uniform probability distributions.

The rollout policy eventually fails when the grid size is increased beyond a certain range and the approximations lose their effectiveness. Note that dividing the grid into smaller sub-grids, then searching these sub-grids instead, generally does not help and will usually lead to more than the minimum number of measurements. This is due to redundant measurements from overlapping subproblems at the sub-grid boundaries. Dynamic programming specifically takes advantage of these overlapping subproblems to achieve optimality. However, the approximations used in approximate dynamic programming can be improved systematically: either by increasing the lookahead in the rollout algorithm, or by including additional steps of policy evaluation and policy improvement to go beyond rollout. Improvements to approximations are expected to extend planning to larger and larger grid sizes.

\section{Dynamic Programming for Gaussian Processes} \label{sec5}
Gaussian processes \citep{rasmussen,gibbs,mackay} are widely used for active sensing of environmental phenomena. Here, we give a variant of our DP algorithm that is applicable to Gaussian processes by specifically considering a robot transect sampling task similar to that described in \citet{cao}. Alternative approaches to dynamic programming applied to Gaussian processes appear in \citet{peters}.

In a robot transect sampling task, a robot changes its position $x'\in\mathbb{R}^2$ continuously while taking sensor measurements at a discrete set of locations $x'_k$ (i.e., one measurement per stage $k$). The outcome of each measurement is determined by the random field $M(x')$, which is a continuous random variable that varies continuously with robot position $x'$. The probability distribution of $M(x')$ is assumed to be a Gaussian process (GP), meaning that any discrete set of outcomes $M(x'_0)=m_0,\ldots,M(x'_k)=m_k$ has a Gaussian probability density; and a corresponding covariance matrix. In order to derive a DP algorithm that is useful for GPs, it is therefore necessary to go beyond the assumption of independent measurement outcomes. Fortunately, this can be done with a simple modification to Algorithm \ref{alg3} using the method of \emph{{state augmentation}} \citep{bert}. 

A GP model for regression allows us to predict the location of where the next measurement should be taken, given all previous measurement locations. In this case, the probability density of measurement outcome $m_k$ is:
\begin{equation}
p_k(m_k|m_{0:k-1})={\cal{N}}(m_k, \mu_k, \sigma_k^2),
\label{gp_prob}
\end{equation}
where
\begin{align}
\mu_k&={\bf{p}}^T{\bf{C}}^{-1}{\bf{m}}_{k-1}, \label{first}\\
\sigma_k^2&=\kappa-{\bf{p}}^T{\bf{C}}^{-1}{\bf{p}}.\label{second}
\end{align}
In Equation (\ref{gp_prob}), we use the notation $m_{0:k-1}=m_0,\ldots,m_{k-1}$ to denote the sequence of all previous measurement outcomes, which are also used to form the column vector \linebreak${\bf{m}}_{k-1}=(m_0,\ldots,m_{k-1})^T$ in Equation (\ref{first}). The covariance matrix ${\bf{C}}$ depends on the covariance function $C(x,y)$ of the GP, and has elements $C_{ij}=C(x^{\prime}_i,x^{\prime}_j)+\sigma_\nu^2\delta_{ij}$; where indices $i$ and $j$ run from $0$ to $k-1$, and $\sigma_\nu^2$ is the noise variance. The parameter $\kappa$ is given by $\kappa = C(x^\prime_k,x^\prime_k)$, and the column vector ${\bf{p}}$ has elements $p_i=C(x^\prime_i,x^\prime_k)$; where ${\bf{p}}$ is a column vector, and its transpose ${\bf{p}}^T$ is a row vector. The differential entropy of the GP is given by $H_k(M_k|\sigma_k^2)=H_k(M_k|x'_{0:k})$, and depends only on past measurement locations but not on past measurement outcomes.

The DP recurrence relation can now be derived with reference to Equation (\ref{dpagent}). Assumptions equivalent to \citet{cao} include only one choice of measurement at each stage, and robot dynamics that is deterministic. These assumptions reduce Equation (\ref{dpagent}) to Equation (\ref{dpmesagent}), as in the case of ``Find the Submarine''. Further, $x_k$ and $f_k$ play no role in this model, and so Equation (\ref{dpmesagent}) further reduces to:
\begin{equation}
J_k(x'_k)=
\underset{m_k}{\mathbb{E}}\Big\{h_k(u_k(x'_k),m_k)\Big\}+\underset{u'_k\in U'_k(x'_k)}{\operatorname{max}}\ J_{k+1}(v_k(x'_k,u'_k)).\label{notright}
\end{equation}
However, this recurrence relation is not quite right because we assumed independence to derive it. In particular, the entropy (the first term on the right-hand side) should be replaced by $H_k(M_k|x'_{0:k})$ from our GP model. This means the state given by $x'_k$ in Equation (\ref{notright}) is no longer sufficient and must now be augmented by $x^\prime_{0:k-1}$ to give the new state $x'_{0:k-1},x'_k=x'_{0:k}$. Therefore, the size of the state-space has increased substantially, and approximate DP methods like the one given below will generally be required. The corresponding DP recurrence relation is now written as 
\begin{equation}
J_k(x'_{0:k})=H_k(M_k|x'_{0:k})+\underset{u'_k\in U'_k(x'_k)}{\operatorname{max}}\ J_{k+1}(x'_{0:k}, v_k(x'_k,u'_k)).
\label{gpdp}
\end{equation}
The DP algorithm corresponding to Equation (\ref{gpdp}) now takes into account all past measurement locations $x^\prime_{0:k-1}$ leading to $x'_k$ so that the entropy at stage $k$ may be found. At stage $k$, the robot then chooses control $u'_k$ to reach the most informative measurement location $x'_{k+1}$ at the next stage. This leads to $x'_{0:k}, v_k=x'_{0:k+1}$ for the argument of $J_{k+1}$.

The DP recurrence relation given by Equation (\ref{gpdp}) is expected to give similar results when used in the rollout algorithm to the ``approximate maximum entropy path planning'' presented in \citet{cao}. However, the strength of our DP framework is that it is more general, and therefore can be used to describe more diverse situations. For example, if the robot dynamics is stochastic instead of deterministic, we can simply appeal to Equation~(\ref{dpagent}) to get the following DP recurrence relation:
\begin{equation}
J_k(x'_{0:k})=H_k(M_k|x'_{0:k})+\underset{u'_k\in U'_k(x'_k)}{\operatorname{max}}\ \underset{w_k}{\mathbb{E}}\Big\{J_{k+1}(x'_{0:k}, v_k(x'_k,u'_k, w_k))\Big\}.
\end{equation}
Alternatively, instead of considering a single random field, we might be interested in sensing several random fields simultaneously; such as the salinity and temperature of a water body. We then have more than one choice of sensor measurement available at each stage. Again, appealing to Equation~(\ref{dpagent}), we might choose to model this using the following DP recurrence relation:
\begin{equation}
J_k(x'_{0:k})=\underset{u_k\in U_k(x'_k)}{\operatorname{max}}\ \sum_i u_k^i H_k^i(M_k^i|x'_{0:k})+\underset{u'_k\in U'_k(x'_k)}{\operatorname{max}}\ J_{k+1}(x'_{0:k}, v_k(x'_k,u'_k)),\label{multiple}
\end{equation}
where $M_k^1(x')$ and $M_k^2(x')$ might be the salinity and temperature fields, for example. In this case, possible measurement choices at each stage would include $u_k=(1, 0)$, \linebreak$u_k=(0,1)$, or $u_k=(1,1)$. Presumably, the default case is the measurement $u_k=(1,1)$ where both salinity and temperature are measured simultaneously at each stage. However, in some~circumstances there may be hard constraints on either the number of measurements possible at each stage, or the type of measurement that can be taken at each stage. This could be due to constraints on power consumption, storage of samples, sensor response times, etc. The DP recurrence relation given by Equation (\ref{multiple}) is able to properly account for these types of measurement constraints, as well as any kinematic constraints on the robot or vehicle. This is done through the constraint sets $U_k(x'_k)$ and $U'_k(x'_k)$, which depend on the robot position $x'_k$ at time $k$. These are just three examples, but other possibilities for DP recurrence relations can also be derived from Equation (\ref{dpagent}) under different modelling assumptions.

A modified version of Algorithm \ref{alg3} is now proposed for solving a GP. Specifically, Lines 7--12 in Algorithm \ref{alg3} are replaced with Lines 7--13 in Algorithm \ref{dpalg5}. The main change is the extra assignment on Line 8, which is necessary for prediction of the $i$th stage entropy, $H_i(M_i|\sigma_i^2)$. On Line 13, the entropy predictions from stages $k$ to $N-1$ are then added together and stored. An additional evaluation of $\sigma_{k}^2$ following Line 2 in Algorithm \ref{alg3} is also required in order to define $H_k(M_k|x'_{0:k})$ on Line 13. Further slight modifications of Algorithm \ref{alg3} may also be required, depending on the precise form of the DP recurrence relation considered. The assignment on Line 8 requires computation of ${\bf{C}}^{-1}$, which takes ${\cal{O}}(i^3)$ time using exact matrix inversion. However, a tighter bound of ${\cal{O}}((i-j)^3)$ may be possible by recognizing that dependencies might only be appreciable between a subset of past locations $x'_{j:i-1}$, rather than all past locations $x'_{0:i-1}$: potentially leading to a much smaller matrix for ${\bf{C}}$. The size of this subset will depend on the length-scale hyperparameters in the covariance function, as well as the distance between each sampling location (this tighter bound will not be realized in one-dimensional GP models where past sampling locations might be re-visited at later stages). In the best case, we can hope to gain a small constant-time overhead with each iteration, and the modified Algorithm \ref{alg3} still scales as ${\cal{O}}(N^2C)$ in the deterministic case. If not, a further reduction in computational time is possible by replacing exact matrix inversion with one of the approximation methods discussed in \citet{rasmussen}. 

\newcommand{\setalglineno}[1]{%
  \setcounter{ALG@line}{\numexpr#1-1}}
\begin{algorithm}
\caption{Modified Stochastic Rollout for GPs}\label{dpalg5}
\begin{algorithmic}[1]
\setalglineno{7}
\For{$i=k+1$ \textbf{to} $N-1$}
\vspace{4pt}
\setalglineno{8}
\State $\sigma_i^2(x'_{0:i})=\kappa-{\bf{p}}^T{\bf{C}}^{-1}{\bf{p}}$
\vspace{6pt}
\setalglineno{9}
\State $\hat{\mu}'_{i}(x'_i)\gets\verb|Generate_base_policy|(x'_{i})$
\vspace{4pt}
\setalglineno{10}
\State $w_i\sim p_{W_i}$
\vspace{4pt}
\setalglineno{11}
\State $x'_{i+1}\gets v_i(x'_i,\hat{\mu}'_i(x'_i),w_i)$
\vspace{6pt}
\setalglineno{12}
\EndFor
\vspace{2pt}
\setalglineno{13}
\State \texttt{Store:} $H_k(M_k|x'_{0:k})+\tilde{J}_{k+1}(x'_{k+1})$
\end{algorithmic}
\end{algorithm}

\section{Conclusions}\label{sec6}
The outcome of this work was the development of a general-purpose dynamic programming algorithm for finding an optimal sequence of informative measurements when complete state information is available (i.e., measurement outcomes are not noisy). This algorithm unifies the design of informative measurements with efficient path planning for robots and autonomous agents. While greedy methods are still the most common approach for finding informative measurements, we showed that an essential characteristic of some types of optimal measurement sequences includes planning for delayed information gains. This seems especially true for path planning in artificial intelligence and robotics, where finding optimal sequences of informative measurements is more likely to lead to a combinatorial optimization problem. We demonstrated a simple path planning problem involving a deterministic agent that could not be solved efficiently using a greedy method. We also showed an approximate dynamic programming solution to this problem that clearly exhibited delayed information gains due to planning trade-offs taking place. 

An obvious application of the proposed algorithm is to make efficient use of sensors on an autonomous robot or vehicle that is exploring a new environment. Some of the difficulties of applying dynamic programming and reinforcement learning to robotics are outlined in the review by \citet{peters2}. A major strength of our dynamic programming algorithm is that it can simultaneously take into account sensor constraints and kinematic constraints of robots and autonomous vehicles. Continuous states, controls, and hard constraints can also be handled using a version of the rollout algorithm called model predictive control (MPC). This requires having a map of the environment, as well as an accurate method for localizing the position of the robot or vehicle on the map. Another application of our~algorithm is efficient active sensing of spatially continuous phenomena via a Gaussian process model. We showed how to include different types of sensor constraints while simultaneously including the dynamics and kinematic constraints of the sensor platform. This application will be explored further in a future paper. 
\subsection*{Acknowledgements}
PL thanks Fritz Sommer for informative discussions on this topic.
\section*{Appendix 1: Objective Maximized by Extended DP Algorithm}
Define the maximum entropy of the $(N-k)$-stage problem to be
\begin{equation}
J^*_k(x'_k,x_k)=\underset{\pi^{\prime k}}{\operatorname{max}}\ \mathbb{E'}\left\{
\underset{\pi^k}{\operatorname{max}}\ \mathbb{E}\left\{\sum_{i=k}^{N-1}h_i(x_i,\mu_i(x'_i,x_i),m_i)+h_N(x'_N,x_N)\right\}\right\},
\end{equation}
where $\pi^{\prime k}=\{\mu'_k,\ldots,\mu'_{N-1}\}$, $\pi^k=\{\mu_k,\ldots,\mu_{N-1}\}$, and where $\mathbb{E}'$ is an expectation over $w_k,\ldots,w_{N-1}$, and $\mathbb{E}$ is an expectation over $m_k,\ldots,m_{N-1}$.

\noindent \textbf{Proposition:} 
The maximum entropy of the $N$-stage problem, $J^*_0(x'_0,x_0)$, is equal to $J_0(x'_0,x_0)$, given by the last step of a DP algorithm that starts with the terminal condition: $J_N(x'_N,x_N)$\linebreak$=h_N(x'_N,x_N)$, and proceeds backwards in time by evaluating the following recurrence relation:
\begin{align}
J_k(x'_k,x_k)=
\underset{u'_k\in U'_k(x'_k)}{\operatorname{max}}\ \underset{w_k}{\mathbb{E}}\Big\{
\underset{u_k\in U_k(x'_k,x_k)}{\operatorname{max}}\ \underset{m_k}{\mathbb{E}}\Big\{h_k(x_k,u_k,m_k)\nonumber\\
+J_{k+1}(v_k(x'_k,u'_k,w_k), f_k(x_k,u_k,m_k))\Big\}\Big\},
\end{align}
from stage $k=N-1$ to stage $k=0$. 

\noindent\textbf{Proof:} 
We give a proof by mathematical induction. At stage $k=N$, we have $J^*_N(x'_N,x_N)=h_N(x'_N,x_N)=J_N(x'_N,x_N)$, proving the base case to be true. 

Now assume for some $k\leq N-1$, and all $(x'_{k+1},x_{k+1})$, that $J^*_{k+1}(x'_{k+1},x_{k+1})=J_{k+1}(x'_{k+1},x_{k+1})$. We need to show that $J^*_{k}(x'_{k},x_{k})=J_{k}(x'_{k},x_{k})$ to complete the proof. We have,
\begin{align*}
J^*_k(x'_k,x_k)&=\underset{(\mu'_k,\pi^{\prime {k+1}})}{\operatorname{max}}\ \mathbb{E'}\Big\{
\underset{(\mu_k,\pi^{k+1})}{\operatorname{max}}\ \mathbb{E}\Big\{h_k(x_k,\mu_k(x'_k,x_k),m_k)\nonumber\\
&\hspace{15pt}+\sum_{i=k+1}^{N-1}h_i(x_i,\mu_i(x'_i,x_i),m_i)+h_N(x'_N,x_N)\Big\}\Big\},\\
&=\underset{\mu'_k}{\operatorname{max}}\ 
\underset{w_k}{\mathbb{E}}\Big\{
\underset{\mu_k}{\operatorname{max}}\ \underset{m_k}{\mathbb{E}}\Big\{h_k(x_k,\mu_k(x'_k,x_k),m_k)
\nonumber\\
&\hspace{15pt}+\underset{\pi^{\prime {k+1}}}{\operatorname{max}}\ \underset{\{w_{k+1},..\}}{\mathbb{E}}\Big\{
\underset{\pi^{k+1}}{\operatorname{max}}\ \underset{\{m_{k+1},..\}}{\mathbb{E}}\Big\{\sum_{i=k+1}^{N-1}h_i(x_i,\mu_i(x'_i,x_i),m_i)+h_N(x'_N,x_N)\Big\}\Big\}\Big\}\Big\},\\
&=\underset{\mu'_k}{\operatorname{max}}\ 
\underset{w_k}{\mathbb{E}}\Big\{
\underset{\mu_k}{\operatorname{max}}\ \underset{m_k}{\mathbb{E}}\Big\{h_k(x_k,\mu_k(x'_k,x_k),m_k)+J^*_{k+1}(x'_{k+1},x_{k+1})\Big\}\Big\},\\
&=\underset{\mu'_k}{\operatorname{max}}\ 
\underset{w_k}{\mathbb{E}}\Big\{
\underset{\mu_k}{\operatorname{max}}\ \underset{m_k}{\mathbb{E}}\Big\{h_k(x_k,\mu_k(x'_k,x_k),m_k)+J_{k+1}(x'_{k+1},x_{k+1})\Big\}\Big\},\\
&=\underset{u'_k\in U'_k(x'_k)}{\operatorname{max}}\ \underset{w_k}{\mathbb{E}}\Big\{
\underset{u_k\in U_k(x'_k,x_k)}{\operatorname{max}}\ \underset{m_k}{\mathbb{E}}\Big\{h_k(x_k,u_k,m_k)\nonumber\\
&\hspace{15pt}+J_{k+1}(v_k(x'_k,u'_k,w_k), f_k(x_k,u_k,m_k))\Big\}\Big\},\\
&=J_k(x'_k,x_k).
\end{align*}
In the first equation above, we used the definition of $J^*_k$, $\pi^{\prime k}$, and $\pi^k$. In the second equation, we interchanged the maxima over $\pi^{\prime k+1}$ and $\pi^{k+1}$ with $h_k$ because the tail portion of an optimal policy is optimal for the tail subproblem. We also used the linearity of expectation. In the third equation, we used the definition of $J^*_{k+1}$, and in the fourth equation we used the inductive hypothesis: $J^*_{k+1}=J_{k+1}$. In the fifth equation, we substituted $v_k$ and $f_k$ for $x'_{k+1}$ and $x_{k+1}$, and $u'_k$ and $u_k$ for $\mu'_k$ and $\mu_k$. In the sixth equation, we used the definition of~$J_k$. \hfill $\blacksquare$

\section*{Appendix 2: Guess My Number}

In this example, an integer is selected uniformly at random in the range $[0,n-1]$, and the problem is to find the minimum number of yes/no questions guaranteed to determine this integer. The bisection method and binary search algorithm are efficient methods for solving this type of problem, even with noisy responses \citep{waeber}. Here, we use DP to demonstrate the optimality of these methods for maximizing the entropy of a set of binary comparisons represented by the yes/no questions. 

Let $X$ be the unknown random integer between 0 and $n-1$. Given a proper subinterval of consecutive integers between 0 and $n-1$, let $M$ return ``yes'' or ``no'' to the question: ``Is the unknown integer within this subinterval?''. We also make the following definitions:
\begin{align*}
x_k &= \text{size of integer range considered at stage } k,\\
u_k &= \text{size of proper subinterval containing the unknown integer at stage } k,
\end{align*}
and the following parameterizations:
\begin{align*}
p_k(m_k|x_k,u_k) &= 
\begin{cases}
u_k/x_k & m_k= \text{``yes''},\\
(x_k-u_k)/x_k & m_k = \text{``no''},
\end{cases}\\ \\
f_k(x_k,u_k,m_k) &= 
\begin{cases}
u_k & m_k= \text{``yes''},\\
x_k-u_k & m_k = \text{``no''}.
\end{cases} 
\end{align*}
With these definitions and parameterizations, Equations (\ref{dp}) and (\ref{infocontent}) lead to the following DP recurrence relation:
\begin{align}
J_k(x_k)=\underset{u_k\in U^+(x_k)}{\operatorname{max}}\left\{\frac{u_k}{x_k}\left (\log_2{\frac{x_k}{u_k}}+J_{k+1}\left (u_k\right )\right )+\frac{x_k-u_k}{x_k}\left (\log_2{\frac{x_k}{x_k-u_k}}+J_{k+1}\left (x_k-u_k\right )\right )\right\},
\end{align}
where $U^{+}(x_k)$ is the set $\{1,2,\ldots,x_k-1\}$.

The DP algorithm starts with the terminal condition $J_N(1)=0$ bits, as before. Considering the tail subproblem for measurement $N-1$ and $x_{N-1}=2$, leads to
\begin{align*}
J_{N-1}(2)&=\frac{1}{2}(\log_2{2}+J_{N}\left (1\right ))+\frac{1}{2}(\log_2{2}+J_{N}\left (1\right )),\\
&= 1\text{ bit.} \ \ \ (u^*_{N-1}=1)
\end{align*}
Now the subproblem for $x_k=3$ requires the value for $J_{N-1}(2)$, which only becomes an overlapping subproblem if we move to measurement $N-2$:
\begin{align*}
J_{N-2}(3)&=\operatorname{max}\left\{\frac{1}{3}\left (\log_2{3}+J_{N-1}\left (1\right )\right )+\frac{2}{3}\left (\log_2{\frac{3}{2}}+J_{N-1}\left (2\right )\right ),\right.\nonumber\\
&\left.\hspace{50pt} \frac{2}{3}\left (\log_2{\frac{3}{2}}+J_{N-1}\left (2\right )\right )+\frac{1}{3}\left (\log_2{3}+J_{N-1}\left (1\right )\right )
\right \},\nonumber\\
&= \log_2{3}\text{ bits.} \ \ \ (u^*_{N-2}=1\text{ or }u^*_{N-2}=2)
\end{align*}
This solution tells us that if we start with three integers ($x_{N-2}=3$) and choose the first subinterval to be length 1 ($u^*_{N-2}=1$), then with probability $1/3$ we can determine the unknown integer with one question. Otherwise, two questions will be necessary. Now consider the DP for $x_{N-2}=4$,
\begin{align*}
J_{N-2}(4)&=\operatorname{max}\left\{\frac{1}{4}\left (\log_2{4}+J_{N-1}\left (1\right )\right )+\frac{3}{4}\left (\log_2{\frac{4}{3}}+J_{N-1}\left (3\right )\right ),\right.\nonumber\\
&\hspace{50pt}\left. \frac{1}{2}\left (\log_2{2}+J_{N-1}\left (2\right )\right )+\frac{1}{2}\left (\log_2{2}+J_{N-1}\left (2\right )\right ), \right.\\
\end{align*}
\begin{align*}
&\hspace{60pt}\left. \frac{3}{4}\left (\log_2{\frac{4}{3}}+J_{N-1}\left (3\right )\right ) + \frac{1}{4}\left (\log_2{4}+J_{N-1}\left (1\right )\right )
\right \},\\\nonumber
&= 2\text{ bits.} \ \ \ (u^*_{N-2}=2)
\end{align*}
In the first equation, $J_{N-1}(3)$ is replaced with $\log_2{3}-2/3$ to get the final result because $J_{N-1}(3)$ cannot be fully resolved in a single measurement. This result can be derived in a similar way to $J_{N-2}(3)$, but instead, using $J_N(2)=0$. From the solution for $J_{N-2}(4)$, it is seen that two bits of information can be gained from two binary questions provided each subinterval divides the previous subinterval in half: $u^*_{N-2}=2$ when $x_{N-2}=4$, and $u^*_{N-1}=1$ when $x_{N-1}=2$. This solution can be continued, giving an upper bound on the entropy as $1$ bit per question for $M$. Then $\log_2{n}$ bits for $X$ means the unknown random integer is guaranteed to be found after a number of yes/no questions equal to $\lceil \log_2{n}\rceil$. This result demonstrates that methods such as the binary search algorithm and the bisection method maximize the entropy of a set of binary comparisons.

\bibliographystyle{plainnat}
\bibliography{Loxley_references} 

\end{document}